%% file: main.tex
\documentclass[10pt,twocolumn,letterpaper]{article}

\usepackage[pagenumbers]{cvpr} % To force page numbers, e.g. for an arXiv version

\usepackage{graphicx}
\usepackage{amsmath}
\usepackage{amssymb}
\usepackage{booktabs}

\usepackage{enumitem}

\usepackage{hyperref}
\hypersetup{breaklinks,colorlinks}

\usepackage[capitalize]{cleveref}
\crefname{section}{Sec.}{Secs.}
\Crefname{section}{Section}{Sections}
\Crefname{table}{Table}{Tables}
\crefname{table}{Tab.}{Tabs.}

\def\cvprPaperID{906} % *** Enter the CVPR Paper ID here
\def\confName{CVPR}
\def\confYear{2023}

\begin{document}

\title{StyleRes: Transforming the Residuals for Real Image Editing with StyleGAN}

\author{Hamza Pehlivan  \qquad
Yusuf Dalva  \qquad
Aysegul Dundar \\
Bilkent University\\
\tt\small \{hamza.pehlivan,yusuf.dalva\}@bilkent.edu.tr\\
\tt\small adundar@cs.bilkent.edu.tr\\}

\maketitle

\begin{abstract}
We present a novel image inversion framework and a training pipeline to achieve high-fidelity image inversion with high-quality attribute editing. 
Inverting real images into StyleGAN's latent space is an extensively studied problem, yet the trade-off between the image reconstruction fidelity and image editing quality remains an open challenge.
The low-rate latent spaces are limited in their expressiveness power for high-fidelity reconstruction. On the other hand, high-rate latent spaces result in degradation in editing quality.
In this work, to achieve high-fidelity inversion, we learn residual features in higher latent codes that lower latent codes were not able to encode. This enables preserving image details in reconstruction. 
To achieve high-quality editing, we learn how to transform the residual features for adapting to manipulations in latent codes.
We train the framework to extract residual features and transform them via a novel architecture pipeline and cycle consistency losses.
We run extensive experiments and compare our method with state-of-the-art inversion methods. Qualitative metrics and visual comparisons show significant improvements. Code: \href{https://github.com/hamzapehlivan/StyleRes}{https://github.com/hamzapehlivan/StyleRes}

\end{abstract}

\input{1_introduction}

\input{2_relatedwork}

\input{3_method}

\input{4_experiments}
\input{5_conclusion}

\section*{Acknowledgement}

This work has been funded by The Scientific and Technological Research Council of Turkey (TUBITAK), 3501 Research Project under Grant
No 121E097. 
 A. Dundar was supported by Marie Skłodowska-Curie Individual Fellowship.

{\small
\bibliographystyle{ieee_fullname}
\bibliography{egbib}
}

\input{6_supp_release}

\end{document}

%% file: 1_introduction.tex
\section{Introduction}

Generative Adversarial Networks (GANs) achieve high quality synthesis of various objects that are hard to distinguish from real images ~\cite{goodfellow2014generative, karras2019style, zhang2019self, karras2020analyzing, yu2021dual}.
These networks also have an important property that they organize their latent space in a semantically meaningful way; as such, via latent editing, one can manipulate an attribute of a generated image.
This property makes GANs a promising technology for image attribute editing and not only for generated images but also for real images.
However, for real images, one also needs to find the corresponding latent code that will generate the particular real image.
For this purpose, different GAN inversion methods are proposed, aiming to project real images to pretrained GAN latent space \cite{shen2020interpreting,voynov2020unsupervised,harkonen2020ganspace,shen2021closed,wu2021stylespace}. 
Even though this is an extensively studied problem with significant progress, the trade-off between image reconstruction fidelity and image editing quality remains an open challenge.

\newcommand{\interpfigi}[1]{\includegraphics[trim=0 0 0cm 0, clip, height=2.5cm]{#1}}
\begin{figure}
\centering
\scalebox{0.71}{
\addtolength{\tabcolsep}{-5pt}   
\begin{tabular}{ccccc}
\\
\rotatebox{90}{~~~~~~~~~Input} &
\interpfigi{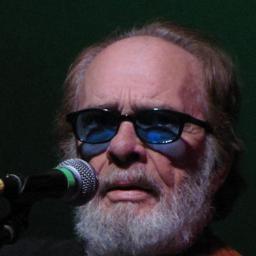} &
\interpfigi{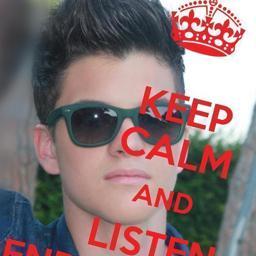} &
\interpfigi{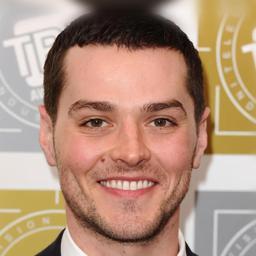} &
\interpfigi{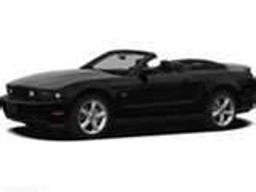} \\
\rotatebox{90}{~~~~~~~~~~e4e \cite{tov2021designing}} &
\interpfigi{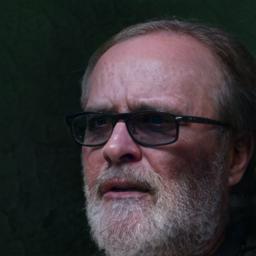} &
\interpfigi{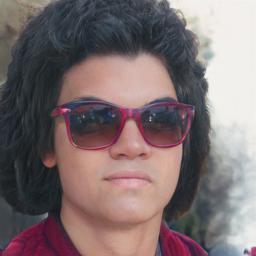} &
\interpfigi{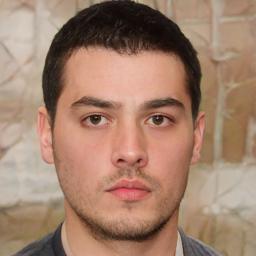} &
\interpfigi{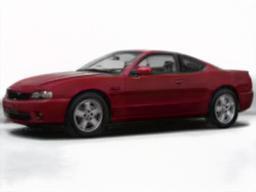} \\
\rotatebox{90}{~~~~~~HFGI \cite{wang2022high}} &
\interpfigi{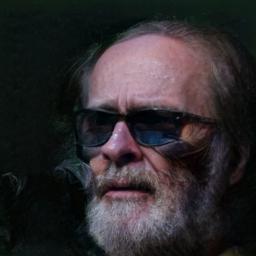}&
\interpfigi{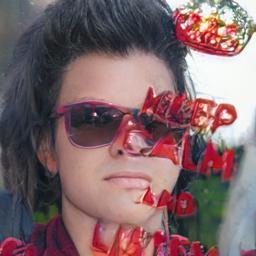} &
\interpfigi{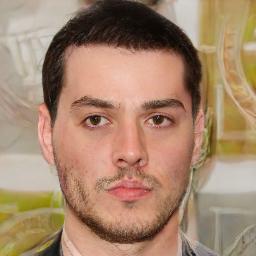} &
\interpfigi{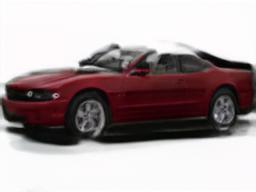} \\
\rotatebox{90}{~~~HyperStyle \cite{alaluf2022hyperstyle}} &
\interpfigi{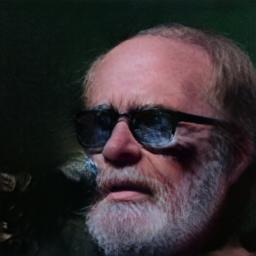} &
\interpfigi{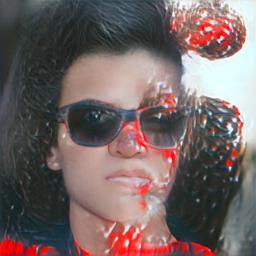} &
\interpfigi{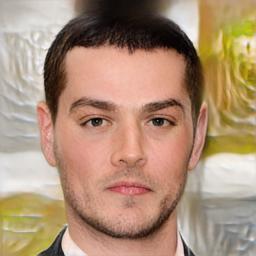} &
\interpfigi{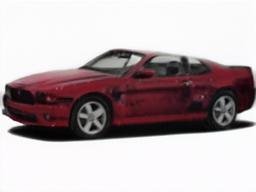} \\
\rotatebox{90}{~StyleRes (Ours)} &
\interpfigi{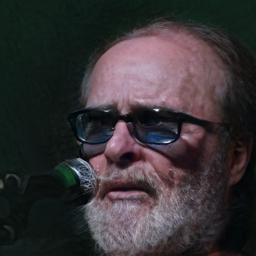} &
\interpfigi{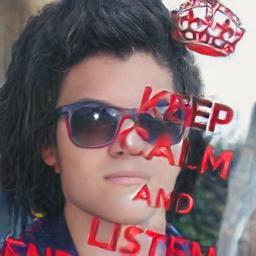} &
\interpfigi{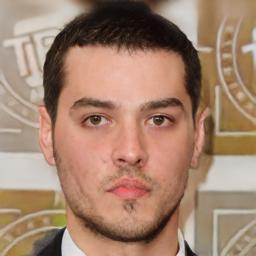} &
\interpfigi{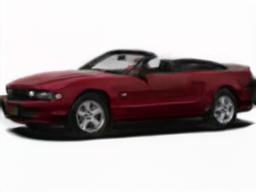} \\
\end{tabular}}
\caption{Comparison of our method with e4e, HFGI, and HyperStyle for the pose, bob cut hairstyle, smile removal, and color change edits. Our method achieves high fidelity to the input and high quality edits.}
\label{fig:teaser}
\end{figure}

The trade-off between image reconstruction fidelity and image editing quality is referred to as the \textit{distortion-editability } trade-off \cite{tov2021designing}.
Both are essential for real image editing. However, it is shown that the low-rate latent spaces are limited in their expressiveness power, and not every image can be inverted with high fidelity reconstruction \cite{abdal2019image2stylegan, richardson2021encoding, tov2021designing, wang2022high}.
For that reason, higher bit encodings and more expressive style spaces are explored for image inversion \cite{abdal2019image2stylegan, abdal2020image2stylegan++}. Although with these techniques, images can be reconstructed with better fidelity, the editing quality decreases since there is no guarantee that projected codes will  naturally lie in the generator's  latent manifold.

% \begin{figure}[t]
% \centering
% \includegraphics[width=1\linewidth]{Figures/img1.pdf}
% \caption{Comparison of our method with HFGI  inversion and editing (smile removal, pose, age reduction, grass). Our method achieves high quality edits   }
% \label{fig:teaser}
% \end{figure}

In this work, we propose a framework that achieves high fidelity input reconstruction and significantly improved editability compared to the state-of-the-art. 
We learn residual features in higher-rate latent codes that are missing in the reconstruction of  encoded features.
This enables us to reconstruct image details and background information which are difficult to reconstruct via low rate latent encodings. 
Our architecture is single stage and learns the residuals based on the encoded features from the encoder and generated features from the pretrained GANs.
We also learn a module to transform the higher-latent codes if needed based on the generated features (e.g. when the low-rate latent codes are manipulated).
This way, when low-rate latent codes are edited for attribute manipulation, the decoded features can adapt to the edits to reconstruct details.
While the attributes are not edited, the encoder can be trained with image reconstruction and adversarial losses.
On the other hand, when the image is edited, we cannot use image reconstruction loss to regularize the network to preserve the details.
To guide the network to learn correct transformations based on the generated features, we train the model with adversarial loss and cycle consistency constraint; that is, after we edit the latent code and generate an image, we reverse the edit and aim at reconstructing the original image.
Since we do not want our method to be limited to predefined edits, during training, we simulate edits by randomly interpolating them with sampled latent codes.

The closest to our approach is HFGI \cite{wang2022high}, which also learns higher-rate encodings.
Our framework is different as we learn a single-stage architecture designed to learn features that are missing from low-rate encodings and we learn how to transform them based on edits.
As shown in Fig. \ref{fig:teaser}, 
our framework achieves significantly better results than HFGI and other methods in editing quality.
In summary, our main contributions are:
\begin{itemize}[leftmargin=*]
    \item We propose a single-stage framework that achieves high-fidelity input embedding and editing.
    Our framework achieves that with  a novel encoder architecture.
    \item We propose to guide the image projection with cycle consistency and adversarial losses. We edit encoded images by taking a step toward a randomly sampled latent code. We expect to reconstruct the original image when the edit is reversed. This way, edited images preserve the details of the input image, and edits become high quality.
    \item We conduct extensive experiments to show the effectiveness of our framework and achieve significant improvements over state-of-the-art for both reconstruction and real image attribute manipulations.

\end{itemize}

%% file: 2_relatedwork.tex
\section{Related Work}

\textbf{GAN Inversion.} GAN inversion methods aim at projecting a real image into GANs embedding so that from the embedding, GAN can generate the given image. Currently, the biggest motivation of the inversion is the ability to edit the image via semantically rich disentangled features of GANs; therefore models aim for high reconstruction and editing quality.
Inversion methods can be categorized into two; 1) methods that directly optimize the latent vector to minimize the reconstruction error between the output and target image \cite{creswell2018inverting, abdal2019image2stylegan, abdal2020image2stylegan++, karras2020analyzing, roich2022pivotal},  2) methods that learn encoders to  reconstruct images over training images \cite{zhu2020domain, tov2021designing, wang2022high}.
Optimization based methods require per-image optimization and iterative passes on GANs, which take several minutes per image.
Additionally, overfitting on a single image results in latent codes that do not lie in GAN's natural latent distribution, leading to poor editing quality.
Among these methods, PTI \cite{roich2022pivotal} takes a different approach, and instead of searching for a latent code that will reconstruct the image most accurately, PTI fine-tunes the generator in order to insert encoded latent code into well-behaved regions of the latent space.
It shows better editability; however, the method still suffers from long run-time optimizations and tunings.
Encoder based methods leverage the knowledge of training sets while projecting images. 
They output results with less fidelity to the input, but edits on the projected latents are better quality. They also operate mostly in real-time. 
They either project the latent code with a single feed-forward pass \cite{zhu2020domain, richardson2021encoding,  tov2021designing}, or two stage feed-forward passes where the first encoder learns to embed an image, the second learns to reconstruct the missing details \cite{wang2022high}, or iterative forward passes where the network tries to minimize the reconstruction loss at each pass by taking the original image and reconstructed output as inputs \cite{alaluf2021restyle, alaluf2022hyperstyle}.
In this work, we propose a novel encoder architecture that achieves significantly better results than state-of-the-art via a single feed-forward pass on the input image. 

\begin{figure*}[t]
\centering
\includegraphics[width=1\linewidth]{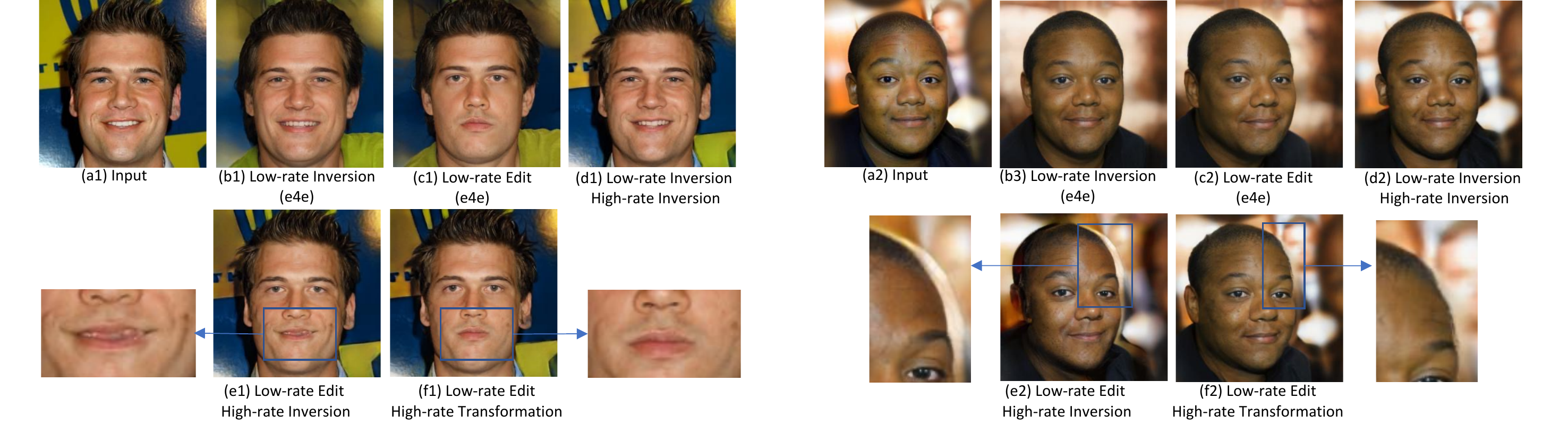}
\caption{When images are encoded to W+ space, as shown in (b), the reconstructions miss many image details. However, the edits are in good quality (c).
Additional features can be learned in higher-rate latent codes. For example, skip connections from encoder to generator in higher resolution features can enable high-fidelity reconstruction to input images as shown in (d).
However, if the high-rate features do not transform with the edits, they result in ghosting effects as shown in (e). In this work, we propose high-rate encoding and transformation for successful inversions and edits (f).}
\label{fig:motivation}
\end{figure*}

\textbf{Image Translation.} There is quite an interest in image translation algorithms that can change an attribute of an image while preserving a given content \cite{wang2018high, dundar2020panoptic, liu2022partial}, especially  for faces editing \cite{starganv2,shen2017learning,xiao2018elegant,zhang2018generative,li2021image,wu2019relgan,gao2021high,hou2022guidedstyle,abdal2021styleflow, dalva2022vecgan}.
These algorithms set an encoder-decoder architecture and train the models with reconstruction and GAN losses \cite{huang2018multimodal,starganv2, zhu2017multimodal, li2021image,yang2021l2m}.
Even though they are successful at image manipulation,
they are trained for a given task and rely on predefined translation datasets.
On the other hand, it is shown that GANs that are trained to synthesize objects in an unconditional way can  be used for attribute manipulation \cite{karras2019style,brock2018large,karras2020analyzing} via their semantically rich disentangled features.
This makes pretrained GANs promising technology for image editing, given that via their ability of image generation, they can achieve many downstream tasks simultaneously. 
Many methods have been proposed for finding latent directions to edit images \cite{shen2020interpreting,voynov2020unsupervised,harkonen2020ganspace,shen2021closed,wu2021stylespace}.
These directions are explored both in supervised \cite{shen2020interpreting, abdal2021styleflow} and unsupervised ways \cite{voynov2020unsupervised,harkonen2020ganspace,shen2021closed,wu2021stylespace} resulting in exploration of attribute manipulations beyond the predefined attributes of labeled datasets.
However, finding these directions is only one part of the problem; to achieve real image editing, there is a need for a successful image inversion method which is the topic of this paper.

%% file: 3_method.tex
\begin{figure*}[t]
\centering
\includegraphics[width=\linewidth]{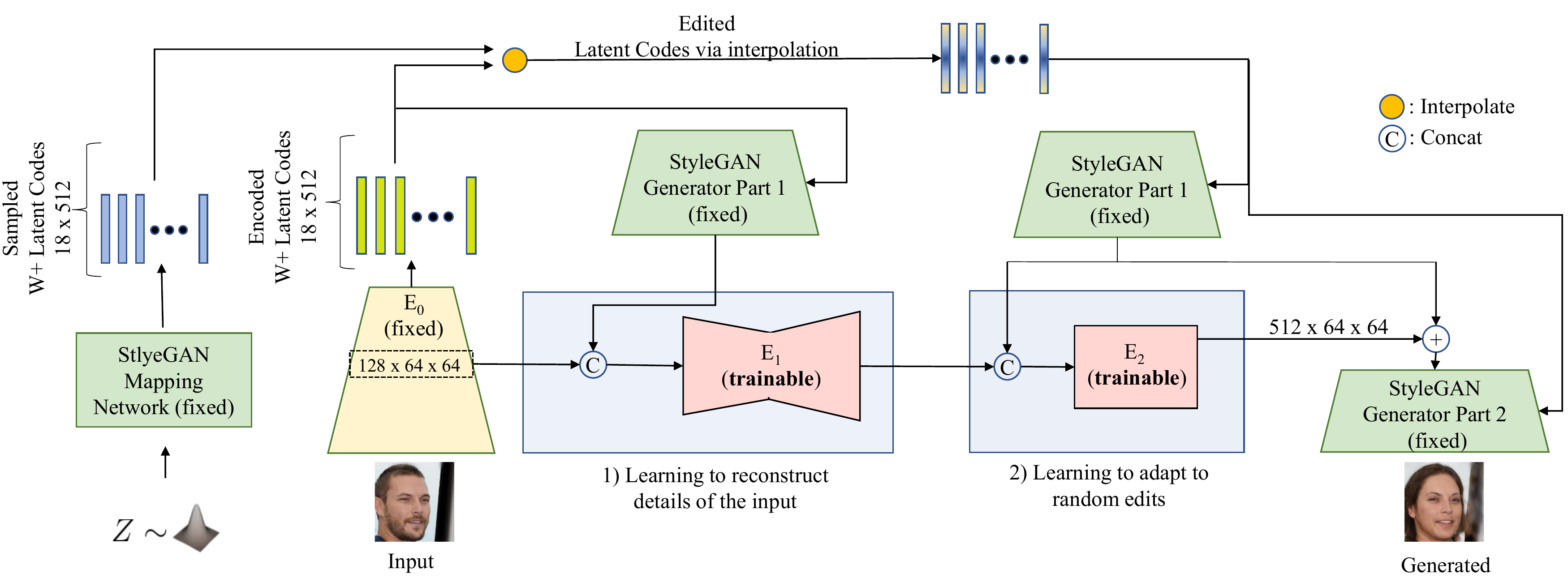}
\caption{StyleRes encodes missing features for high-fidelity reconstruction of given input via the first encoder, $E_1$.
Those encoded features are the ones which could not be encoded to low-rate W+ space  via $E_0$ due to the information bottleneck.
Through the second encoder, $E_2$, StyleRes learns to transform features based on the manipulated features.
During training, latent codes are edited by interpolating encoded W+'s with randomly generated ones by StyleGAN's mapping network. During inference, they are edited with semantically meaningful directions discovered by methods such as InterfaceGAN and GANSpace.
Note that StyleGAN generator is shown as two parts just for the ease of visualizing the diagram. First part includes the layers that generate features to $64\times64$ and the second part generates the higher resolution features and final image.}
\label{fig:architecture}
\end{figure*}

\section{Method}

In Section \ref{sec:mot}, we describe the motivation for our approach. 
The model architecture and the training procedure are presented in Sections \ref{sec:arch} and \ref{sec:training}, respectively.

\subsection{Motivation}
\label{sec:mot}

Current inversion methods aim at learning an encoder to project an input image into  StyleGAN's natural latent space so that the inverted image is editable.
However, it is observed by previous works that when images are encoded  into GAN's natural space (low-rate W or W+ space), their reconstructions suffer from low fidelity to input images, as shown in Fig. \ref{fig:motivation}(b). 
On the other hand, if the image is encoded to higher feature maps of pretrained GANs, they will not be editable since they are not inverted to the semantic space of GANs.
On the other hand, low-rate inversion is editable, as shown in Fig. \ref{fig:motivation}(c). 

StyleGAN, during training and inference, relies on direct noise inputs to generate stochastic details.
These additional noise inputs provide stochastic image details such as for face images, the exact placement of hairs, stubble, freckles, or skin pores which are not modeled by W+ space.
During inversion to W+ space alone, that mechanism is discarded. 
One can tune the noise maps as well via the reconstruction loss; however, then there will be no mechanism to adapt those stochastic details to attribute manipulation.
For example, the freckles should move as the pose is manipulated, but with noise optimization alone, it will not be possible. 

We propose to embed images to both low-rate and high-rate embeddings.
Consistent with the design of StyleGAN, we aim at encoding the overall composition and high-level aspects of the image to W+ space (low-rate encoding).
Our goal is to embed the stochastic image details and the diverse background details, which are difficult to reconstruct from W+ space to higher latent codes. 
However, this setting also requires a mechanism for encoded image details to adopt manipulations in the image.
Otherwise, it will cause a ghosting effect, as shown in Fig \ref{fig:motivation}(e).
For example, in Fig \ref{fig:motivation}(e1), the smile is removed by W+ code edit; however, the higher-rate encodings stay the same and cause artifacts around the mouth area. In Fig. \ref{fig:motivation}(e2), the pose is edited but higher detail encodings did not move and caused blur. 

In this work, we design an encoder architecture and training pipeline to achieve both learning residual features (the ones W+ space could not reconstruct) and how to transform them to be consistent with the manipulations.

\subsection{StyleRes Architecture}
\label{sec:arch}

Our method utilizes an encoder $E_0$ that can embed images into $W+$ latent space and a StyleGAN generator $G$. 
In our setup, we utilize a pretrained encoder for $E_0$ \cite{tov2021designing} and StyleGAN2 generator \cite{karras2020analyzing} and fix them in our training.
Because it is difficult to preserve image details only from low-rate latent codes, we also extract high level features from the encoder. 
First, the high and low rate features are extracted as $F_0, W^+ = E_0(x)$ using the encoder $E_0$ from the input image $x$ as shown in Fig. \ref{fig:architecture}.
$F_0$ has the spatial dimension of $64\times64$ and provides us with more image details.
Next, from $F_0$, our aim is to encode residuals that are missing from the image reconstruction.
For this goal, we set an encoder, $E_1$, which takes $F_0$ as input. 
Since the goal of $E_1$ is to encode residual features, we also feed the generated features with $ W^+$ from the StyleGAN generator,  $G_{W}=G_{0\rightarrow{}n}(W)$, where the arrow operator indicates the indices of convolutional layers used from $G$.

%  The high rate features will be added to generator features $G_{W}=G_{0\rightarrow{}n}(W)$, where the arrow operator indicates the indices of convolutional layers used from $G$. Because $E_0$ is trained only for low rate features, $F_0$ needs further processing. We could further process it with an another high rate encoder $E$ to obtain final features $F$, and add them to $G_{W}$, similar to how HFGI works. However, we argue that separating the features of $E$ and $G$ leads to sub-optimal performance because $F$ cannot adjust itself to incoming $G_{W}$ properly. Therefore, we train an encoder $E_1$ to retrieve an aligned feature map $F_a$, using both encoder and generator feature maps:
\begin{equation}
\label{eq:E1}
 F_a = E_1( F_0, G_{W+})
\end{equation}

With the inputs of $F_0$ and $G_{W+}$, $E_1$ can learn what the missing features are by comparing the encoded features $F_0$  and generated ones, $G_{W+}$, so far. 
While $E_1$ can learn the residual features, it is not guided on how to transform them if images are edited. 
For that purpose, we train $E_2$, which takes $F_a$ and edited features from the generator.
Since we do not target predefined edits (e.g., smile, pose, age), we simulate the edits  by taking random directions. 
Specifically, we sample a $z$ vector from the normal distribution and obtain $W^+_r$ by StyleGAN's mapping network, $M$; $W^+_r=M(z)$.
Next, we take a step towards $W^+_r$ to obtain a mixed style code $W^+_\alpha$. 

\begin{equation}
\label{eq:alpha}
 W^+_\alpha= W^+ + \alpha \frac{W^+_r - W^+}{10}
 \end{equation}

% \[ W_\alpha = W )_i, i > n \]
where 
%$i$ enumerates all of the style vectors, and
$\alpha$ controls the degree of the edit. During the training, we assign $\alpha$ to $0$ (no edit) or a value in the range of $(4, 5)$  with $50\%$ chance.
We adjust $F_a$ to the altered generator features $G_\alpha=G_{0\rightarrow{}n}(W_\alpha)$ using a second encoder $E_2$ and obtain the final feature map F:

\begin{equation}
\label{eq:E2}
F = E_2(F_a, G_\alpha)
\end{equation}

Finally, $F$ and  $G_\alpha$ are summed, and given as an input to next convolutional layers in the generator.  
Architectural details of $E_1$ and $E_2$ are given in Supplementary.

% Throughout the training, we also simulate the possible editings by taking a random step towards to the randomly generated style code. $G$ will take inputs from both low rate $W+$ and high rate $F$ space. The illustration of our design is given in Fig. \ref{fig:architecture}.
%To prevent this, we utilize both randomly sampled latent codes $W_r$. $W_r$ is obtained with a $z$ vector sampled from normal distribution, and StyleGAN's mapping network $M$; $W_r=M(z)$. 

\begin{figure*}[t]
\centering
\includegraphics[width=\linewidth]{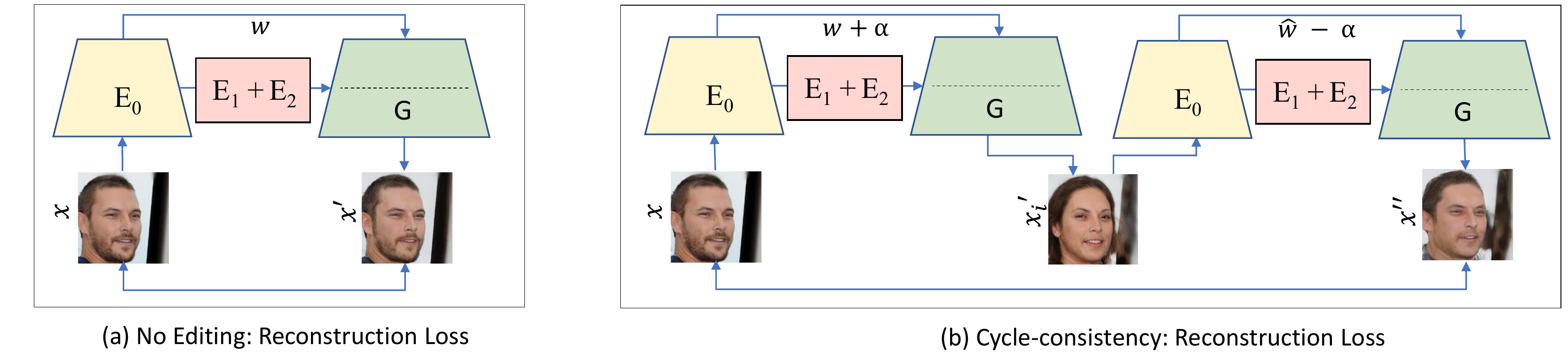}
\caption{We train StyleRes with (a) no editing based reconstruction and (b) cycle consistency based reconstruction losses.
Many details from the inference pipeline are omitted for brevity.
With reconstruction losses, the model learns to preserve the image details.
We additionally apply adversarial losses on $x_i'$; this way, when the image is edited, the transformation network learn to output realistic images.
With cycle consistency based reconstruction, the network is regularized to keep the input details also during edits.
We use additional losses as explained in Section \ref{sec:losses}.}
\label{fig:cycle}
\end{figure*}

\subsection{Training Phases}
\label{sec:training}

To train our model with the capabilities of high-fidelity inversion and high-quality editing, we use two training phases, as shown in Fig. \ref{fig:cycle}.

\textbf{No Editing Path.} In this path, we reconstruct images with no editing on encoded $W^+$. This refers to the case where $\alpha=0$ in Eq. \ref{eq:alpha}.
This is also the training path other inversion methods solely rely on.
While this path can teach the network to reconstruct images faithfully, it does not include any edits and cannot guide the network to high-quality editability.

\textbf{Cycle Translation Path.} In this path, we edit images by setting $\alpha$ a value in the range of $(4, 5)$, which we found to work best in our ablation studies.
Via the edit, the generator outputs image $x_i'$.
Next, we feed this intermediate output image to the encoder and reverse the edit by inverting the addition presented in Eq. \ref{eq:alpha}. The generator reconstructs $x''$, which is supposed to match the input image $x$.
The cycle translation path is important because there is no ground-truth output for the edited image, $x_i'$. Adversarial loss can guide $x_i'$ to look realistic but will not guide the network to keep the input image details if edited.

\subsection{Training Objectives}
\label{sec:losses}
% In our setting, $E_1$ and $E_2$ have trainable parameters and the other modules are fixed.

\textbf{Reconstruction Losses.} For both no editing and cycle consistency path outputs, our goal is to reconstruct the input image. 
To supervise this behavior, we use  $L_2$ loss, perceptual loss, and identity loss between the input and output images.
The first reconstruction loss is as follows, where $x$ is the input image, $x'$ and $x''$ are output images, as given in Fig. \ref{fig:cycle}:

\begin{equation}
    \begin{split}
        \mathcal{L}_{rec-l2} = ||x' - x||_2 + 
        ||x'' - x||_2
    \end{split}
    \label{eqn:rec_loss}
\end{equation}

We use perceptual losses from VGG ($\Phi$) at different feature layers ($j$) between these images from the loss objective as given in Eq. \ref{eq:percep}. 

\begin{equation}
\label{eq:percep}
\small
    \mathcal{L}_{rec-p} = ||\Phi_{j}(x') - \Phi_{j}(x) ||_2 + ||\Phi_{j}(x'') - \Phi_{j}(x) ||_2
\end{equation}

Identity loss is calculated with a pre-trained network $A$. $A$ is an ArcFace model \cite{deng2019arcface} when training on the face domain and a domain specific ResNet-50 model \cite{tov2021designing} for our training on car class. 
\begin{equation}
    L_{rec-id} = (1 - \langle A(x),A(x')\rangle)+(1 - \langle A(x),A(x'')\rangle)
\end{equation}

\textbf{Adversarial Losses.} We also use adversarial losses to guide the network to output images that match the real image distribution. 
This objective improves realistic image inversions and edits. 
We load the pretrained discriminator from StyleGAN training, $D$, and train the discriminator together with the encoders.

\begin{equation}
 \begin{split}
    L_{adv} = 2\log{D(x)} + \log{(1-D(x'))} \\ + \log{(1-D(x'_i))}
    \end{split}
\end{equation}

\textbf{Feature Regularizer.} To prevent our encoder from deviating much from the original StyleGAN space, we regularize the residual features to be small:
\begin{equation}
    L_{F} = \sum_{F \in \phi} \|F\|_2
\end{equation}

\textbf{Full Objective.}
We use the overall objectives given below. The hyper parameters are provided in Supplementary.

% \begin{equation}
%     \min_{\theta_E} \lambda_1 L_2 + \lambda_2 L_{vgg} + \lambda_3 L_{id} - \lambda_4 L_{adv} + \lambda_5 L_{F}
% \end{equation}

\begin{equation}
    \begin{split}
        \underset{E_1,E_2}{\min} \underset{D}{\max} \lambda_{a}\mathcal{L}_{adv} +  \lambda_{r1} \mathcal{L}_{rec-l2} + \lambda_{r2} \mathcal{L}_{rec-p}\\
        +\lambda_{r3} \mathcal{L}_{rec-id}+
        \lambda_{f} \mathcal{L}_{F}
    \end{split}
    \label{eqn:full_loss}
\end{equation}

% 

% The discriminator parameters $\theta_D$ are optimized as
% \begin{equation}
%     \min_{\theta_D} \mathbb{E}[D(\hat{x})] - \mathbb{E}[D(x)] + \frac{\gamma}{2}\mathbb{E} [ \| \nabla D(x)\|_2^2]
% \end{equation}
% , where $\gamma$ controls the amount of gradient regularization. Note that we do not use any of the reconstruction losses because we are simulating an incoming edit in this case. 

%% file: 4_experiments.tex
\newcommand{\interpfigt}[1]{\includegraphics[trim=0 0 0cm 0, clip, width=2.5cm]{#1}}
% comparison images
% smile removal
\begin{figure*}
\centering
\scalebox{0.71}{
\addtolength{\tabcolsep}{-5pt}   
\begin{tabular}{ccccccccccc}
\\
% \rotatebox{90}{~~~~Smile (-)} &
% \interpfigt{Figures/face_imgs/smile_r/input/20151.jpg} &
% \interpfigt{Figures/face_imgs/inversion/e4e/20151.jpg} &
% \interpfigt{Figures/face_imgs/smile_r/e4e/20151.jpg} &
% \interpfigt{Figures/face_imgs/inversion/hfgi/20151.jpg} &
% \interpfigt{Figures/face_imgs/smile_r/hfgi/20151.jpg} &
% \interpfigt{Figures/face_imgs/inversion/hyperstyle/20151.jpg} &
% \interpfigt{Figures/face_imgs/smile_r/hyperstyle/20151.jpg} &
% \interpfigt{Figures/face_imgs/inversion/ours/20151.jpg} &
% \interpfigt{Figures/face_imgs/smile_r/ours/20151.jpg} \\
\rotatebox{90}{~~~~~~~1. Smile (-)} &
\interpfigt{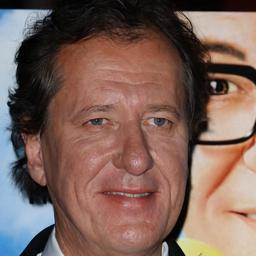} &
\interpfigt{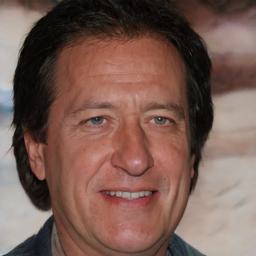} &
\interpfigt{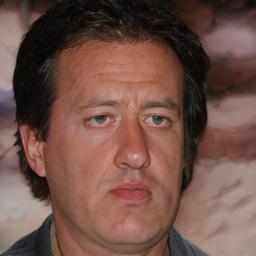} &
\interpfigt{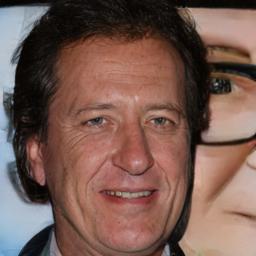} &
\interpfigt{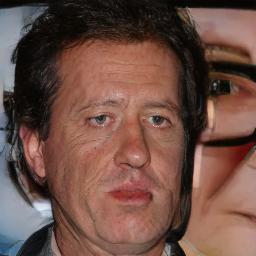} &
\interpfigt{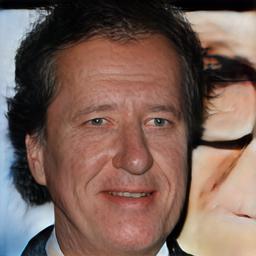} &
\interpfigt{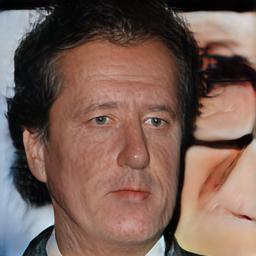} &
\interpfigt{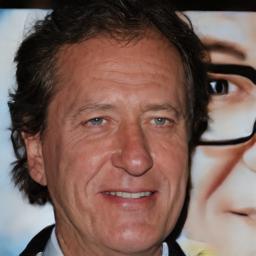} &
\interpfigt{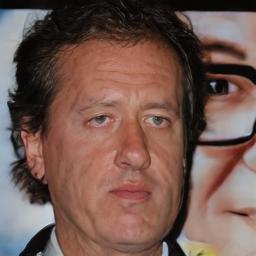} \\
\rotatebox{90}{~~~~~~~2. Age (-)} &
\interpfigt{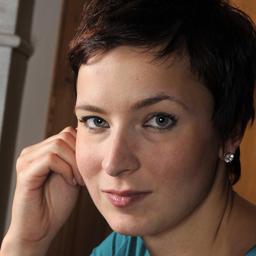} &
\interpfigt{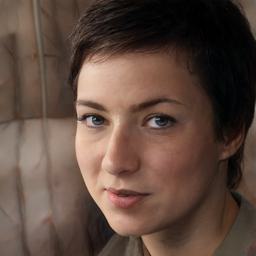} &
\interpfigt{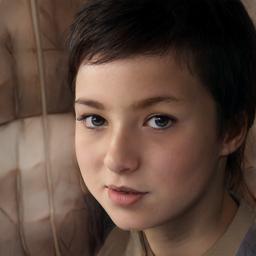} &
\interpfigt{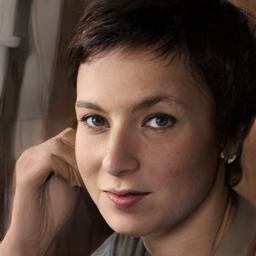} &
\interpfigt{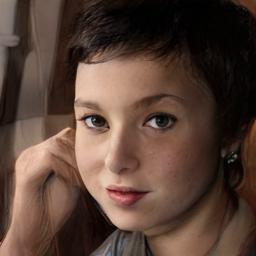} &
\interpfigt{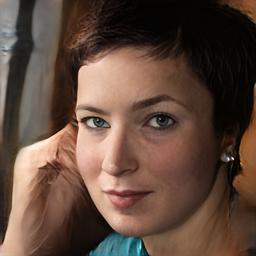} &
\interpfigt{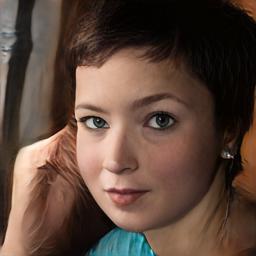} &
\interpfigt{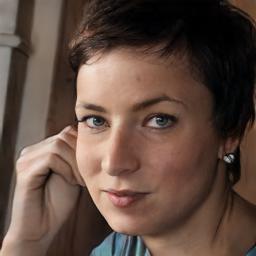} &
\interpfigt{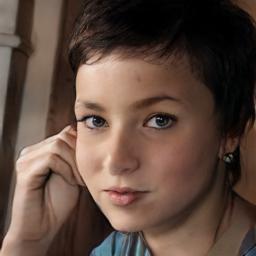}\\
\rotatebox{90}{~~~~~~~~3. Age (+)} &
\interpfigt{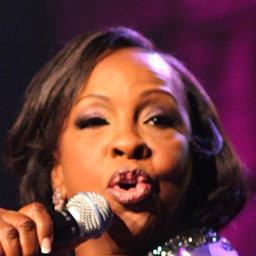}&
\interpfigt{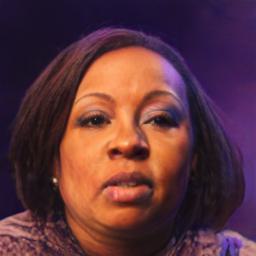} &
\interpfigt{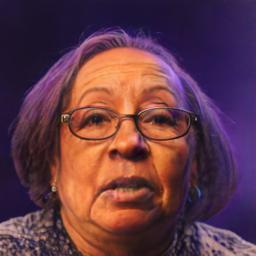} &
\interpfigt{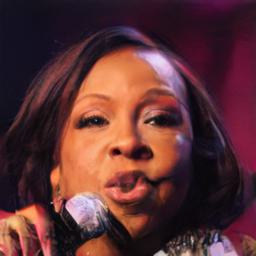} &
\interpfigt{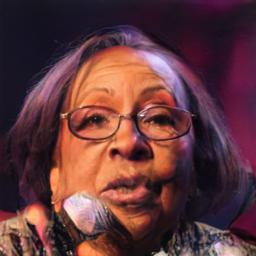} &
\interpfigt{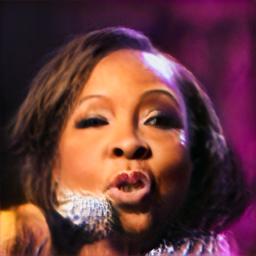} &
\interpfigt{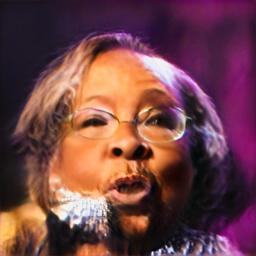} &
\interpfigt{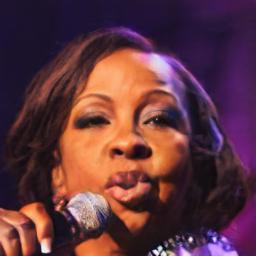} &
\interpfigt{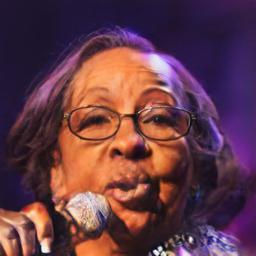} 
\\
\rotatebox{90}{4. Eye Openness} &
\interpfigt{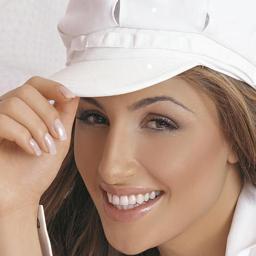} &
\interpfigt{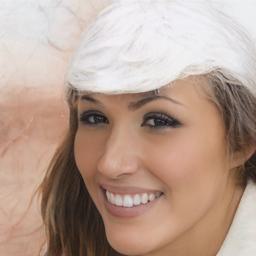} &
\interpfigt{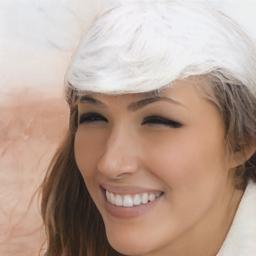} &
\interpfigt{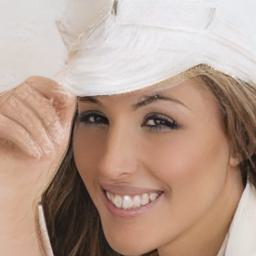} &
\interpfigt{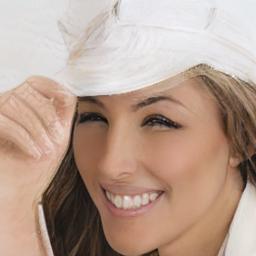} &
\interpfigt{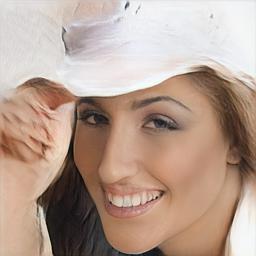} &
\interpfigt{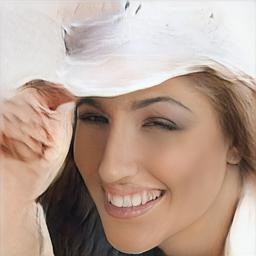} &
\interpfigt{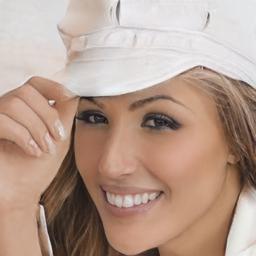} &
\interpfigt{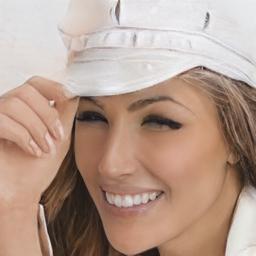} \\
\\
\rotatebox{90}{~~~~~~~5. Pose(-)} &
\interpfigt{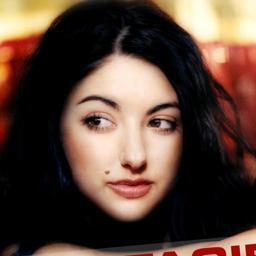} &
\interpfigt{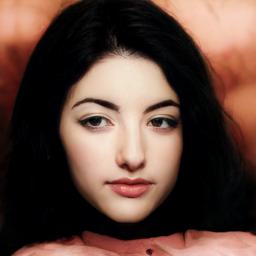} &
\interpfigt{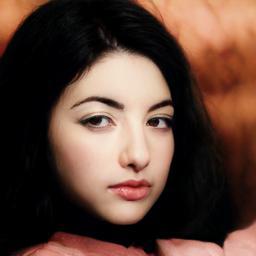} &
\interpfigt{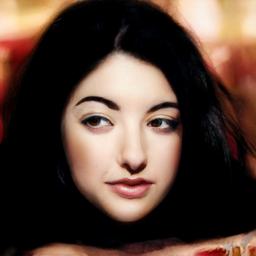} &
\interpfigt{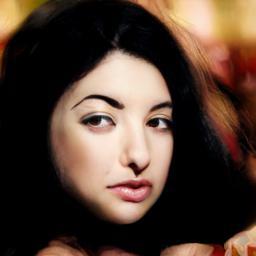} &
\interpfigt{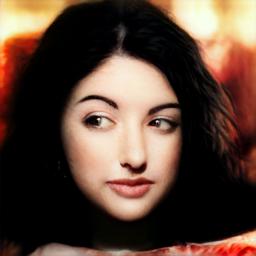} &
\interpfigt{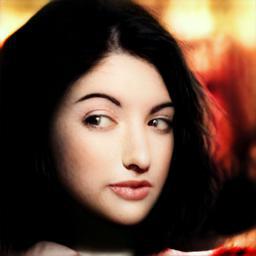} &
\interpfigt{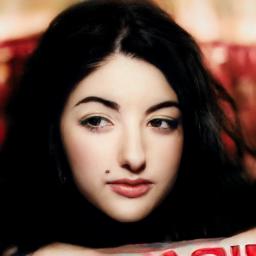} &
\interpfigt{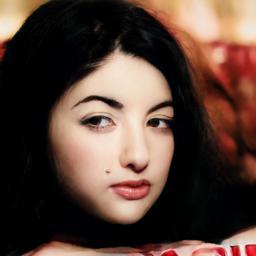}  \\
\rotatebox{90}{~~~~~~~6. Pose(+)} &
\interpfigt{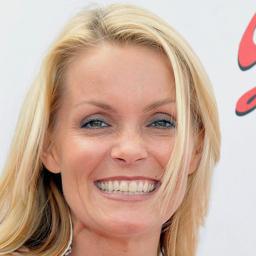} &
\interpfigt{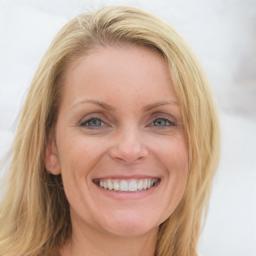} &
\interpfigt{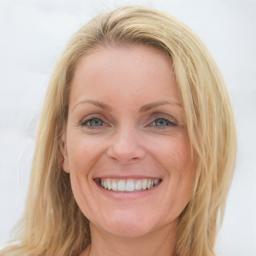} &
\interpfigt{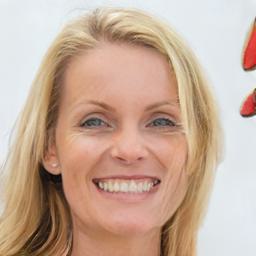} &
\interpfigt{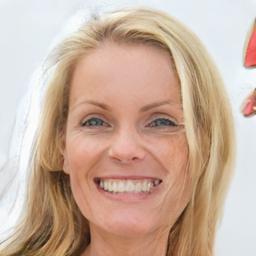} &
\interpfigt{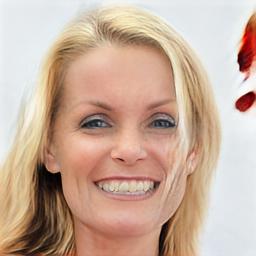} &
\interpfigt{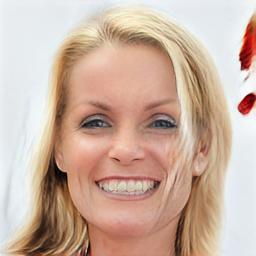} &
\interpfigt{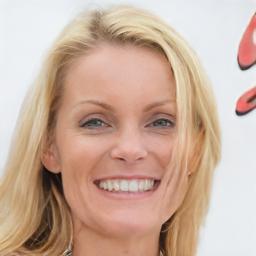} &
\interpfigt{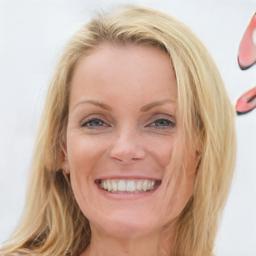}
\\
\rotatebox{90}{~~~~~~~~7. Beard} &
\interpfigt{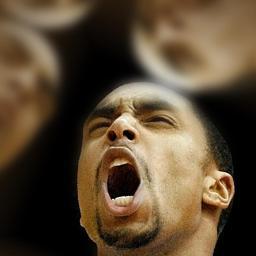}&
\interpfigt{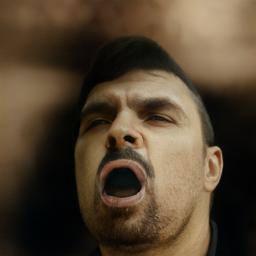} &
\interpfigt{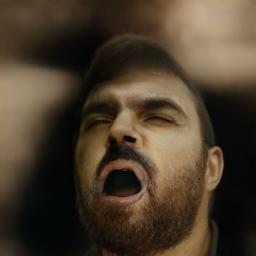} &
\interpfigt{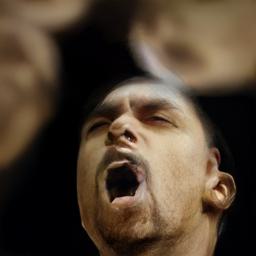} &
\interpfigt{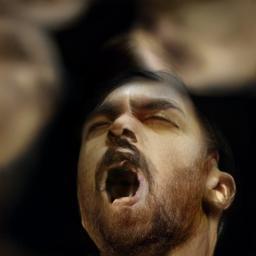} &
\interpfigt{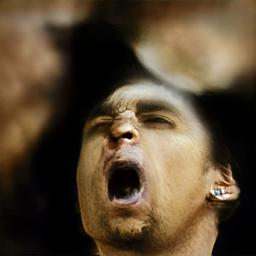} &
\interpfigt{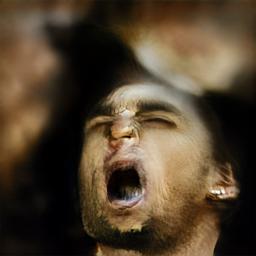} &
\interpfigt{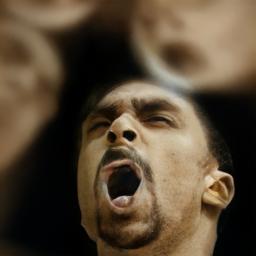} &
\interpfigt{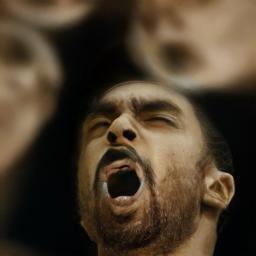} 
\\
\rotatebox{90}{~~~~8. Color} &
\interpfigt{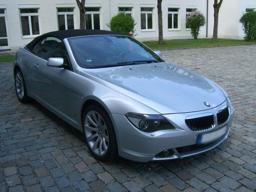} &
\interpfigt{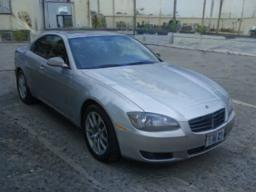} &
\interpfigt{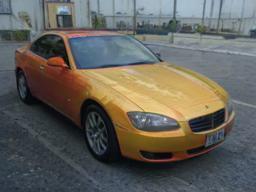} &
\interpfigt{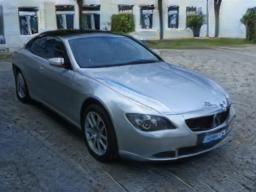} &
\interpfigt{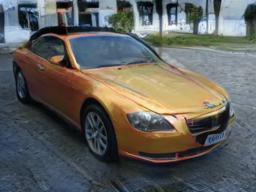} &
\interpfigt{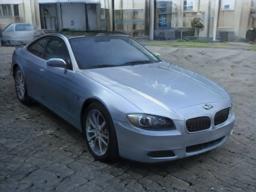} &
\interpfigt{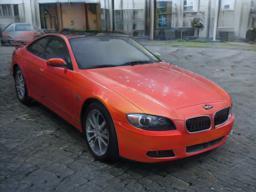} &
\interpfigt{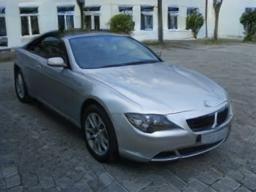} &
\interpfigt{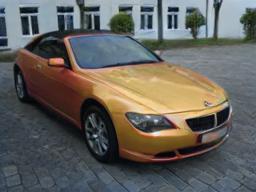}
\\
\rotatebox{90}{~~~~9. Grass} &
\interpfigt{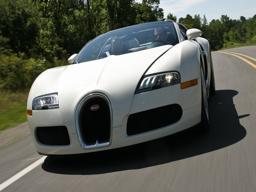} &
\interpfigt{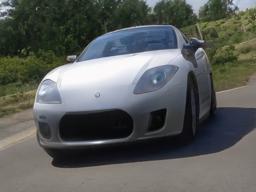} &
\interpfigt{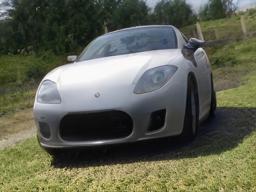} &
\interpfigt{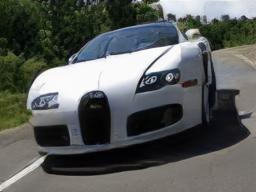} &
\interpfigt{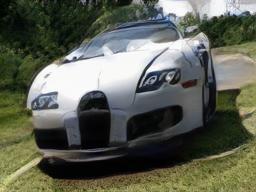} &
\interpfigt{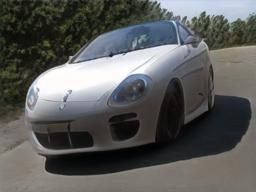} &
\interpfigt{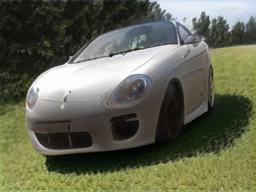} &
\interpfigt{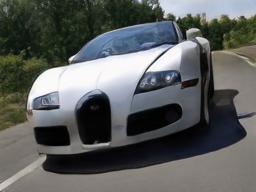} &
\interpfigt{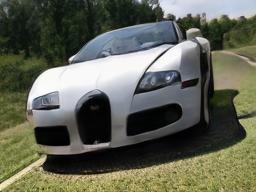}
\\
&  Input &  \multicolumn{2}{c}{e4e \cite{tov2021designing}} &  \multicolumn{2}{c}{HFGI \cite{wang2022high}} &  \multicolumn{2}{c}{HyperStyle \cite{alaluf2022hyperstyle}}  & \multicolumn{2}{c}{StyleRes (Ours)} \\
\end{tabular}
}
\caption{Qualitative results of inversion and editing. For each method, the first column shows inversion, and the second shows InterfaceGAN  \cite{shen2020interpreting} and GANSpace \cite{harkonen2020ganspace} edits.}
\label{fig:results_all}
\end{figure*}

\section{Experiments}

\textbf{Set-up.} For datasets and attribute editing, we follow the previous work  \cite{wang2022high}. For the human face domain, we train the model on the FFHQ \cite{karras2019style} dataset and evaluate it on the CelebA-HQ \cite{karras2017progressive} dataset.
For the car domain, we use Stanford Cars \cite{krause20133d} for training and evaluation.
We run extensive experiments with directions explored with InterfaceGAN \cite{shen2020interpreting}, GANSpace \cite{harkonen2020ganspace}, StyleClip \cite{patashnik2021styleclip}, and GradCtrl \cite{chen2022exploring} methods.

\textbf{Evaluation.}
We report metrics for reconstruction and editing qualities.
For reconstruction, we report Frechet Inception Distance (FID) metric \cite{heusel2017gans}, which looks at the realism by comparing the target image distribution and reconstructed images, Learned Perceptual Image Patch Similarity (LPIPS) \cite{zhang2018unreasonable} and Structural
Similarity Index Measure (SSIM), which compares the target and output pairs at the feature level of a pretrained deep network and pixel level, respectively. 
Additionally, FIDs are calculated for adding and removing the smile attributes on the CelebA-HQ dataset. By using the ground-truth attributes, we add smile to images that do not have smile attribute and calculate FIDs between smile addition edited images and smiling ground-truth images. The same set-up is used for smile removal.
On the Stanford cars dataset, we change grass and color attribute of images, and since there is no ground-truth attribute, we calculate the FIDs between the edited and original images.

% \begin{table*}[t]
% \centering
% \caption{Quantitative results of reconstruction and editing. For reconstruction, we report FID, SSIM, and LPIPS scores. For editing, we report metrics for smile addition and removal. Metrics include FID (+) for smile addition, FID (-) for smile removal, accuracy metrics which uses a classifier to recognize a smile and identity score which uses Arcface model.}
% \begin{tabular}{|l|l|l|l|l|l|l|}
% \hline
% & \multicolumn{3}{c}{Reconstruction} & \multicolumn{2}{c}{Editing}  \\
% \hline
% \textbf{Method} & FID   $\Downarrow$  & SSIM   $\Uparrow$  & LPIPS$\Downarrow$ &  FID (+) $\Downarrow$ & FID (-) \\
% \hline
% e4e& 30.22 & 0.71 & 0.21 & 38.58 & 39.68   \\
% hyperstyle& 16.08 & 0.83 &  0.11  &  26.43 & 25.26  \\
% HFGI & 12.17 & 0.85 & 0.13 & 25.22 & 27.10 \\ 
% Ours & \textbf{6.51} & \textbf{0.91} & \textbf{0.08} & \textbf{21.30} & \textbf{20.85}  \\ 
% \hline
% \end{tabular}
% \label{table:results_all}
% \end{table*}

\begin{table}[t]
\centering
\caption{Quantitative results of reconstruction and editing on CelebA-HQ dataset. For reconstruction, we report FID, SSIM, and LPIPS scores. For editing, we report FID metrics for smile addition (+) and removal (-).}
\resizebox{\linewidth}{!}{
\begin{tabular}{|l|l|l|l||l|l|l|l|l|l|}
\hline
& \multicolumn{3}{c||}{Reconstruction} & \multicolumn{2}{c|}{Editing - FIDs}  \\
\hline
\textbf{Method} & FID  & SSIM  & LPIPS &  Smile(+) & Smile(-)  \\
\hline
% InterFaceGAN \cite{shen2020interpreting} & & &  & 22.8 & 23.9  & 97.8& 99.7 \\
% IdInvert \cite{zhu2020domain} \\
pSp \cite{richardson2021encoding} &  23.86 & 0.75 & 0.17 & 32.47 & 34.0\\
e4e \cite{tov2021designing} & 30.22 & 0.71 & 0.21 & 38.58 & 39.68 \\
ReStyle \cite{alaluf2021restyle} & 24.82 & 0.73 & 0.20 & 30.35 & 33.69 \\
HyperStyle \cite{alaluf2022hyperstyle}& 16.08 & 0.83 &  0.11  &  26.43 & 25.26  \\
HFGI \cite{wang2022high} & 12.17 & 0.85 & 0.13 & 25.22 & 27.10 \\
StyleTransformer \cite{Hu_2022_CVPR} & 21.82 & 0.75 & 0.17 & 34.32 & 34.61\\
FeatureStyle \cite{xuyao2022} & 11.33 & 0.90 & 0.10 & 27.20 & 26.15 \\
% diffae & - &- &- & 17.91 & 20.43 & 99.3 & 99.3 & 0.47& 0.45\\
\hline
StyleRes (Ours) & \textbf{7.04} & \textbf{0.90} & \textbf{0.09} & \textbf{23.52} & \textbf{21.80}  \\

\hline
\end{tabular}}
\label{table:results_smile_r}
\end{table}

\begin{table}[t]
\centering
\caption{Quantitative results of reconstruction and editing on the Stanford Cars Dataset. For reconstruction, we report FID, SSIM, and LPIPS scores. For editing, we report FID metrics for grass addition and color change.}
\resizebox{\linewidth}{!}{
\begin{tabular}{|l|l|l|l||l|l|l|l|l|l|}
\hline
& \multicolumn{3}{c||}{Reconstruction} & \multicolumn{2}{c|}{Editing - FIDs}  \\
\hline
\textbf{Method} & FID  & SSIM  & LPIPS &  Grass & Color  \\
\hline
% InterFaceGAN \cite{shen2020interpreting} & & &  & 22.8 & 23.9  & 97.8& 99.7 \\
e4e \cite{tov2021designing} & 14.04 & 0.50 & 0.32 & 18.02 & 29.79 \\
ReStyle \cite{alaluf2021restyle}& 13.38 & 0.57 & 0.30 & 16.01 & 21.34 \\
HyperStyle \cite{alaluf2022hyperstyle}& 11.64 & 0.63 & 0.28 &  17.13 & 26.30 \\
HFGI \cite{wang2022high} & 9.41 & 0.83  & 0.16 & 14.84 & 26.65 \\
StyleTransformer \cite{Hu_2022_CVPR} & 14.01 & 0.57 & 0.28 & 19.47 & 19.94 \\
% diffae & - &- &- & 17.91 & 20.43 & 99.3 & 99.3 & 0.47& 0.45\\
\hline
StyleRes (Ours) &  \textbf{7.60} &  \textbf{0.83} & \textbf{0.14} &  \textbf{10.64} & \textbf{18.86}   \\

\hline
\end{tabular}}
\label{table:results_car}
\end{table}

\textbf{Baselines.} We compare our method with state-of-the-art image inversion methods pSp \cite{richardson2021encoding}, e4e \cite{tov2021designing}, ReStyle \cite{alaluf2021restyle}, HyperStyle \cite{alaluf2022hyperstyle}, HFGI \cite{wang2022high}, StyleTransformer \cite{Hu_2022_CVPR}, and FeatureStyle \cite{xuyao2022}.
We use the author's released models. Hence, for the Stanford car dataset, some methods are omitted from comparisons if the models are not released.
Among those, we only train HFGI for car model with author's released code since we base our main comparisons with it.
% Among these methods, ReStyle and HyperStyle reconstruct an image via iterative refinement while our method is single-stage.
% HFGI and FeatureStyle encode images into higher spatial dimension latent codes as well as to $W^+$ space, similar to ours.

\textbf{Quantitative Results.} 
In Table \ref{table:results_smile_r}, we provide reconstruction end editing scores on the CelebA-HQ dataset. 
Our method achieves better results than all competing methods on all metrics.
Most significantly, we achieve better FID scores on both reconstruction and editing qualities.
While FeatureStyle achieves comparable SSIM and LPIPS scores on reconstruction, the editing FIDs are worse than our model.
As shown in Table \ref{table:results_car}, we achieve significantly better results than previous methods on the Stanford Car dataset as well.
We also compare the runtime of our method and previous methods.
Our method is significantly faster than HyperStyle (0.125 sec vs. 0.439 sec). 
That is because HyperStyle refines its predicted weight offsets gradually via multiple iterations.
Our method is also faster than HFGI (0.125 sec vs. 0.130 sec), in addition to achieving better scores.
We run a single stage network inference, whereas HFGI first generates an image via e4e and StyleGAN and provides the error map to a second architecture.
The table for runtimes is provided in Supplementary.

\begin{figure}
\centering
\scalebox{0.71}{
\addtolength{\tabcolsep}{-5pt}   
\begin{tabular}{ccccc}
\\
\rotatebox{90}{~~~~~~~~~Input} &
\interpfigi{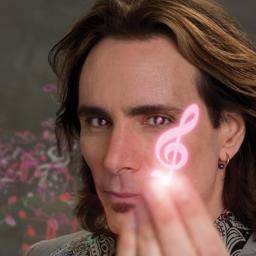}& 
\interpfigi{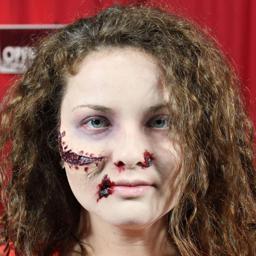} &
\interpfigi{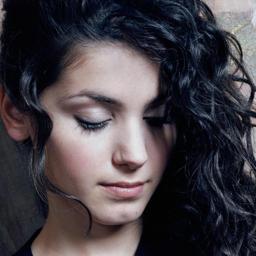} &
\interpfigi{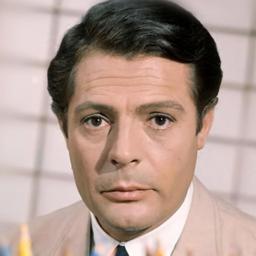} \\
\rotatebox{90}{~~~~~~~~~~e4e \cite{tov2021designing}} &
\interpfigi{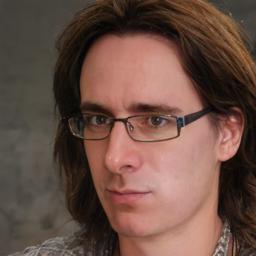} &
\interpfigi{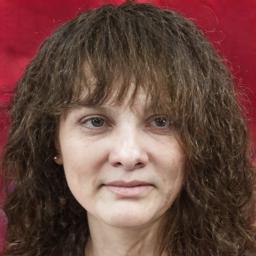} &
\interpfigi{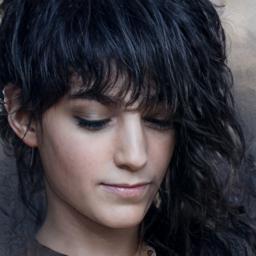} &
\interpfigi{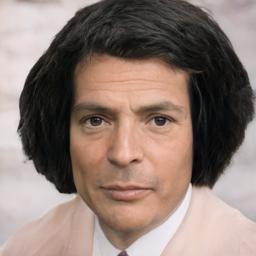} \\
\rotatebox{90}{~~~~~~HFGI \cite{wang2022high}} &
\interpfigi{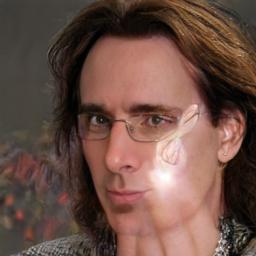}&
\interpfigi{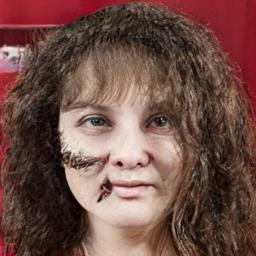} &
\interpfigi{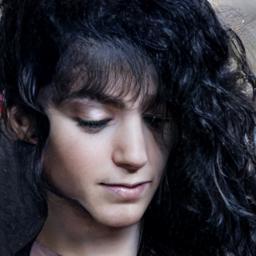} &
\interpfigi{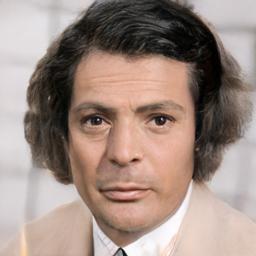} \\
\rotatebox{90}{~~~HyperStyle \cite{alaluf2022hyperstyle}} &
\interpfigi{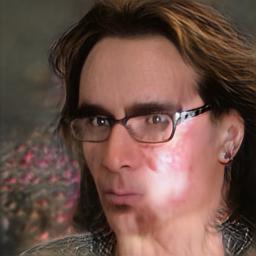} &
\interpfigi{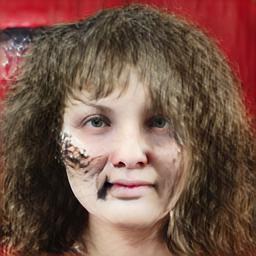} &
\interpfigi{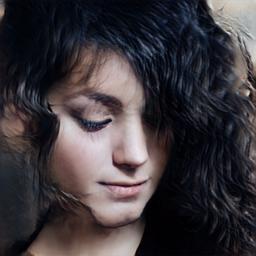} &
\interpfigi{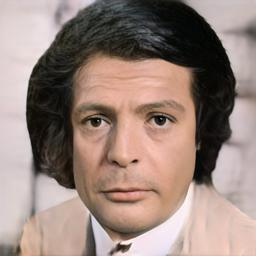} \\
\rotatebox{90}{~StyleRes (Ours)} &
\interpfigi{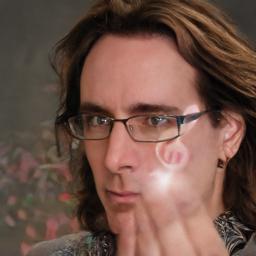} &
\interpfigi{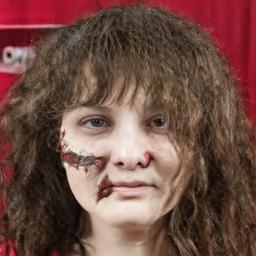} &
\interpfigi{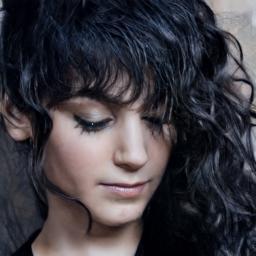} &
\interpfigi{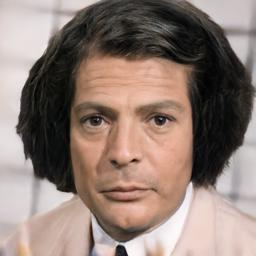} \\
\end{tabular}}
\caption{Comparison of our method with e4e, HFGI, and HyperStyle with StyleClip edits \cite{patashnik2021styleclip}. The first column shows eyeglasses addition, the second and third columns show bangs addition, and the last column shows bob cut hairstyle results.}
\label{fig:styleclip_edits}
\end{figure}

\textbf{Qualitative Results.} We show visuals of inversion and editing results of our method, e4e, HFGI, and HyperStyle in
Fig. \ref{fig:results_all}. 
We provide further comparisons in Supplementary with other attribute editings and with other methods.
Compared to previous methods, our method achieves significantly better fidelity to the input images and preserves the identity and details when edited.
It is the only method in these comparisons that preserves the background, earings (second row), hands (second-fourth rows), and hats (fourth row).
Our method also achieves facial detail reconstruction; for example, in the fifth row, the person has a mole at the corner of her mouth. Among the inversions, our method is the only one preserving that. Furthermore, during the pose edit, it is transformed correctly.
On the seventh row, our method is the only one that achieves the correct inversion and edit.
In car examples, we again achieve high fidelity to the input images both in inversion and editing.
e4e and HyperStyle do not reconstruct the image faithfully. On the other hand, HFGI outputs artefacts during edits.
We additionally show results with edits explored by StyleClip \cite{patashnik2021styleclip} and GradCtrl \cite{chen2022exploring} methods in Figs. \ref{fig:styleclip_edits} and \ref{fig:grad_edits}, respectively.

\newcommand{\interpfigc}[1]{\includegraphics[trim=0 0 0cm 0, clip, height=2.7cm]{#1}}

\begin{figure}
\centering
\scalebox{0.71}{
\addtolength{\tabcolsep}{-5pt}   
\begin{tabular}{ccccc}
\\
\rotatebox{90}{~~~~~~~~~Input} &
\interpfigc{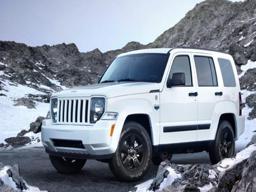} &
\interpfigc{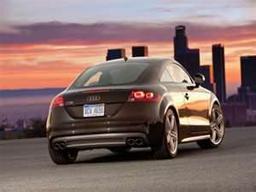} &
\interpfigc{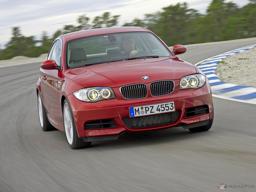}\\
\rotatebox{90}{~~~~~~~~~~e4e \cite{tov2021designing}} &
\interpfigc{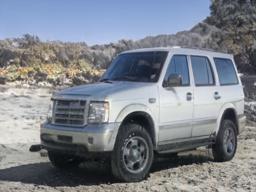}  &
\interpfigc{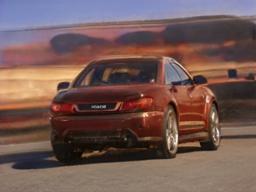}  &
\interpfigc{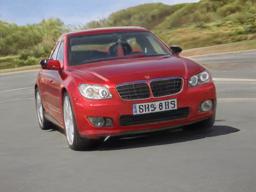}  
\\
\rotatebox{90}{~StyleRes (Ours)} &
\interpfigc{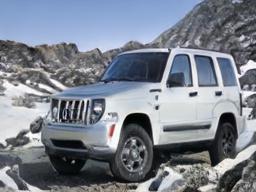}  &
\interpfigc{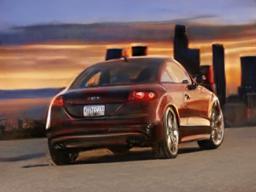}  &
\interpfigc{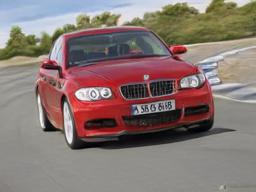}  
 \\
 \rotatebox{90}{~~~~~~~~~Input} &
\interpfigc{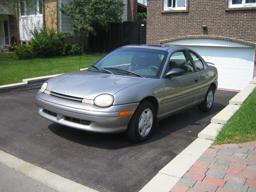} &
% \interpfigc{Figures/car_imgs/tree/input/00403.jpg} &
\interpfigc{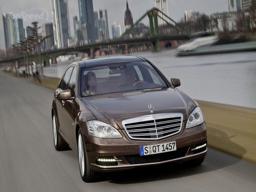} &
\interpfigc{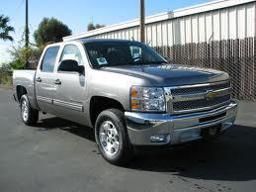} 
\\
\rotatebox{90}{~~~~~~~~~~e4e \cite{tov2021designing}} &
\interpfigc{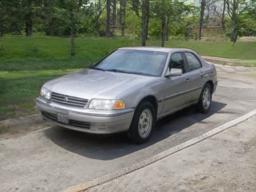}  &
% \interpfigc{Figures/car_imgs/tree/ours/00403.jpg}  &
\interpfigc{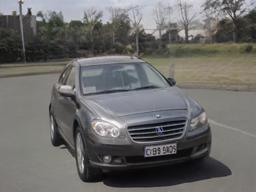}  & 
\interpfigc{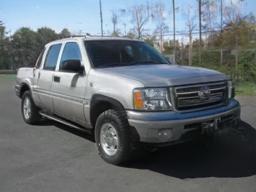}
\\
\rotatebox{90}{~StyleRes (Ours)} &
\interpfigc{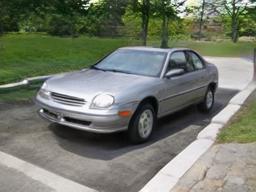}  &
% \interpfigc{Figures/car_imgs/tree/ours/00403.jpg}  &
\interpfigc{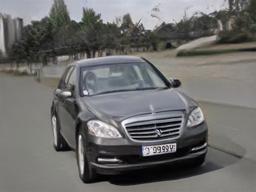}  & 
\interpfigc{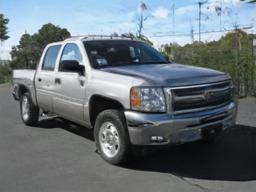}
 \\\end{tabular}}
\caption{Additional results of our method  and e4e with GradCtrl edits \cite{chen2022exploring}.
The first three rows show blue sky edit, and the others show tree background edits.}
\label{fig:grad_edits}
\end{figure}

\begin{figure}
\centering
\scalebox{0.72}{
\addtolength{\tabcolsep}{-5pt}   
\begin{tabular}{ccccc}
\\
\rotatebox{90}{~~~~~~~~~Input} &
\interpfigi{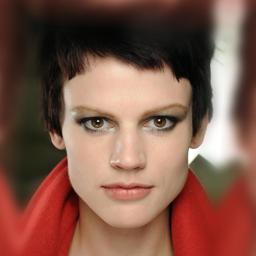}& 
\interpfigi{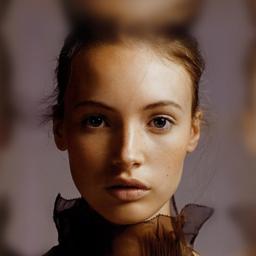}& 
\interpfigi{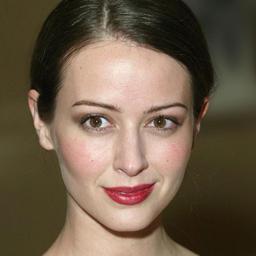}& 
\interpfigi{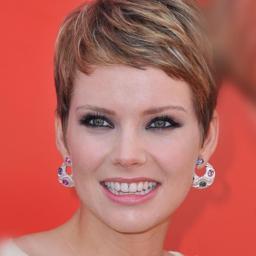} \\
 \rotatebox{90}{~~w/o $E_1$ and $E_2$} &
\interpfigi{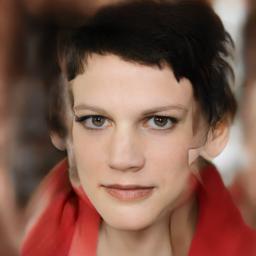} &
\interpfigi{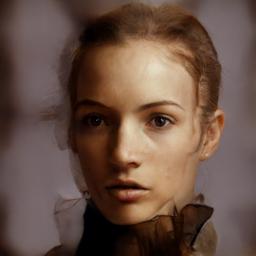} &
\interpfigi{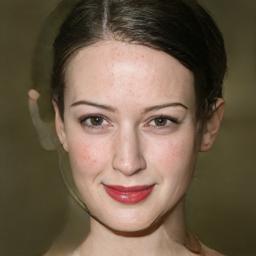} &
\interpfigi{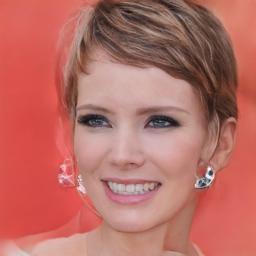} \\
 \rotatebox{90}{~~~~~~~w/o $E_1$} &
\interpfigi{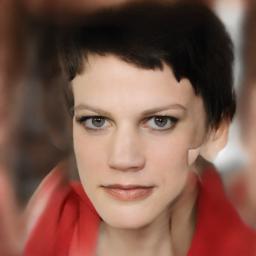} &
\interpfigi{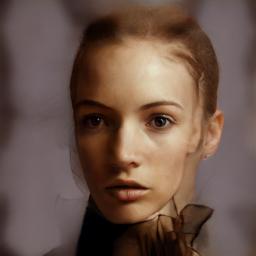} &
\interpfigi{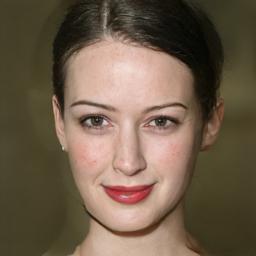} &
\interpfigi{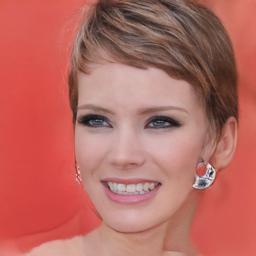} \\
 \rotatebox{90}{~~~~~~~w/o $E_2$} &
\interpfigi{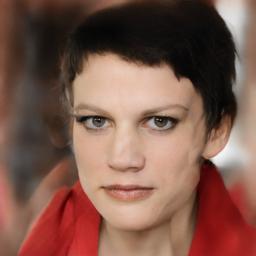} &
\interpfigi{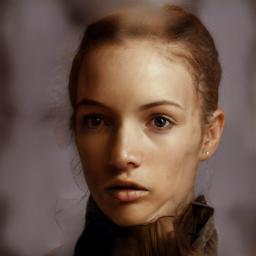} &
\interpfigi{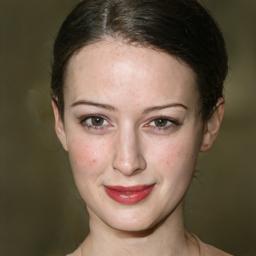} &
\interpfigi{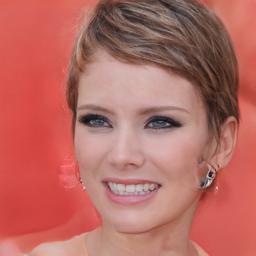} \\
 \rotatebox{90}{~~StyleRes (ours)} &
\interpfigi{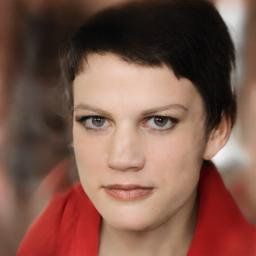} &
\interpfigi{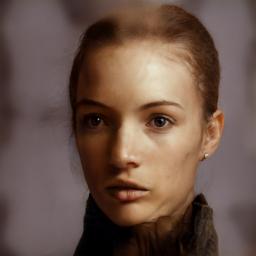} &
\interpfigi{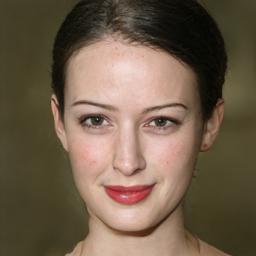} &
\interpfigi{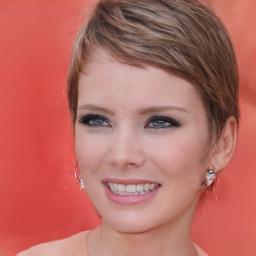} \\
\end{tabular}}
\caption{Ablation Study. Pose edit outputs of our final architecture and architecture with missing modules. w/o $E_1$ or $E_2$, networks struggle to transform features correctly.}
\label{fig:abl_network}
\end{figure}

\begin{figure}[t]
\centering
\includegraphics[width=0.95\linewidth]{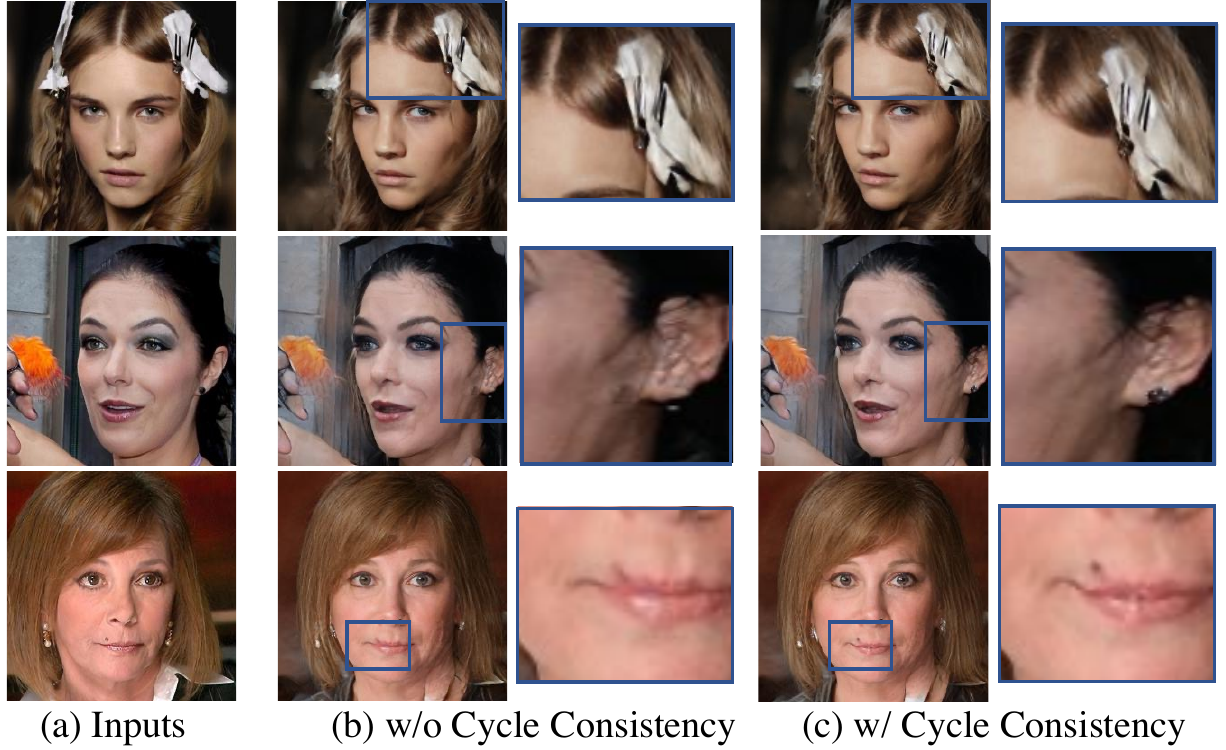}
\caption{Ablation Study. Pose edit outputs of models trained with and without cycle consistency loss. The model trained with cycle consistency is better at preserving the details as shown in enlarged boxes.}
\label{fig:abl_cyc}
\end{figure}

\textbf{Ablation Study.} We run extensive experiments to validate the role of each proposed design choices.
% as given in Table \ref{table:ablation_all}.
% We provide the reconstructions scores and pose edit FIDs.
% For pose edits, we take the first centered 2000 images of CelebA-HQ (yaw between -8 and +8) as our source images and the first 2000 images with yaw between (18-24) as our target distribution images. We edit source images with +2 pose change. 
We first experiment with an architecture that does not have $E_1$ and $E_2$ modules. This experiment refers to the case where we learn a network that does not take input from StyleGAN Part 1 features, neither the original ($G_{W+}$) nor the edited features ($G_{\alpha}$). The network directly takes higher layer features ($F_0$) from the encoder and outputs them to the generator.
As shown in Fig. \ref{fig:abl_network}, the network still tries to learn residual features to achieve reconstructions; however, it achieves poor edits.
Without $E_2$ refers to the experiment with $E_1$ directly outputting features to Generator Part 2 (as shown in Fig. \ref{fig:architecture}).
Without $E_1$ refers to the experiment where $E_2$ directly takes input from the encoder ($F_0$) and ($G_{\alpha}$) as defined in Sec. \ref{sec:arch}.
None of the networks are able to learn the residual features and how to transform them as well as our final architecture since they are not designed to take the original and edited features separately.
%The separation there enables the network to learn which features are missing from the generator part 1 (unedited) and how features should be transformed to adopt to features based on the generated features from generator part 1 (edited).
We also experiment without cycle consistency losses.
Fig. \ref{fig:abl_cyc} shows visual results of methods trained with and without cycle consistency constrain.
We observe that with cycle consistency constrain, our network achieves preserving fine image details even when images are edited.

%% file: 5_conclusion.tex
\section{Conclusion}

We present a novel image inversion framework and a training pipeline to achieve high-fidelity image inversion with high-quality attribute editing. 
In this work, to achieve high-fidelity inversion, we learn residual features in higher latent codes that lower latent codes were not able to encode. This enables preserving image details in reconstruction. 
To achieve high-quality editing, we learn how to transform the residual features for adapting to manipulations in latent codes.
We show state-of-the-art results on a wide range of edits explored with different methods both quantitatively and qualitatively while achieving faster run-time than competing methods.

%% file: 6_supp_release.tex
\appendix

\begin{figure*}[t]
\centering
\includegraphics[width=\linewidth]{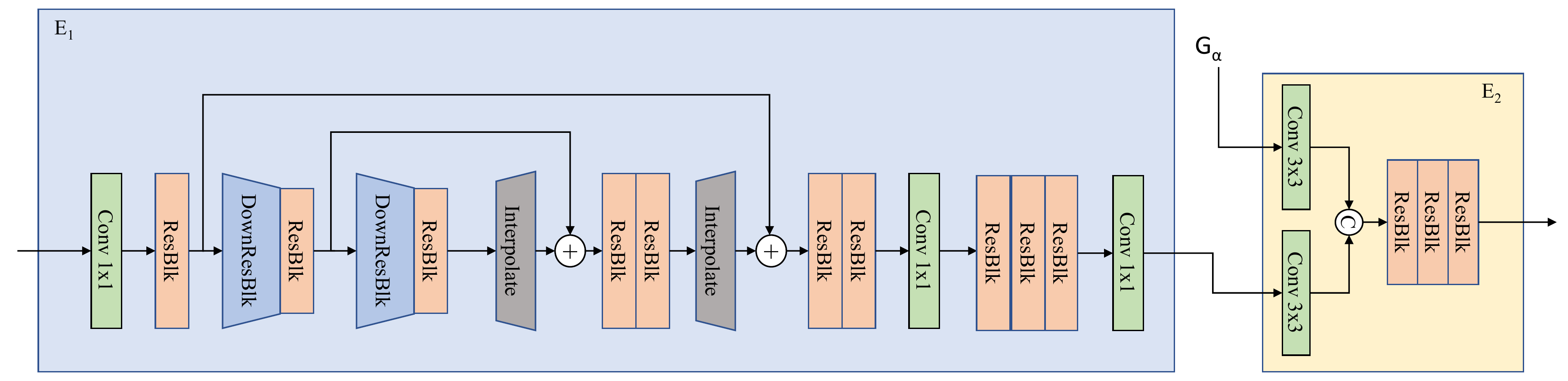}
\caption{Detailed architecture of our encoders. More information regarding dimensions are given in the corresponding text.  }
\label{fig:arch_details}
\end{figure*}

\begin{table}[h]
\centering
\caption{Inference time of the competing methods are given.}
\begin{tabular}{|l|c|}
\hline
\textbf{Method} &  Runtime (sec) \\
\hline
pSp \cite{richardson2021encoding} & 0.088 \\
e4e \cite{tov2021designing} & 0.089 \\
ReStyle \cite{alaluf2021restyle} & 0.365 \\
HyperStyle \cite{alaluf2022hyperstyle} & 0.437 \\
HFGI \cite{wang2022high} & 0.130  \\
StyleTransformer \cite{Hu_2022_CVPR} & 0.063 \\
FeatureStyle \cite{xuyao2022} & 0.526 \\
PTI \cite{roich2022pivotal} & 97.94 \\
\hline
StyleRes (Ours)  & 0.125  \\ 
\hline
\end{tabular}
\label{table:results_runtime}
\end{table}

\begin{table}[t]
\centering
\caption{Quantitative results of reconstruction and editing on CelebA-HQ dataset. We compare with PTI. For reconstruction, we report FID, SSIM, and LPIPS scores. For editing, we report FID metrics for smile addition (+) and removal (-).}
\resizebox{\linewidth}{!}{
\begin{tabular}{|l|l|l|l||l|l|l|l|l|l|}
\hline
& \multicolumn{3}{c||}{Reconstruction} & \multicolumn{2}{c|}{Editing - FIDs}  \\
\hline
\textbf{Method} & FID  & SSIM  & LPIPS &  Smile(+) & Smile(-)  \\
\hline
PTI \cite{roich2022pivotal} & 10.64 & \textbf{0.92} & \textbf{0.07} & 30.29 & 30.11  \\
\hline
StyleRes (Ours) & \textbf{7.04} & 0.90 & 0.09 & \textbf{23.52} & \textbf{21.80}  \\

\hline
\end{tabular}}
\label{table:results_pti}
\end{table}

\section{Training Details}
We use e4e \cite{tov2021designing} as the basic encoder and invert StyleGAN2 \cite{karras2020analyzing} generator. The high level features $F_0$ are the input of the first map2style layer used in e4e. $F_0$ has the spatial dimension of $128\times64\times64$. The intermediate StyleGAN features are extracted as $G_W = G_{0 \rightarrow 8} (W )$, where the arrow operator indicates the indices of convolution layers used. $G_W$ has the spatial dimension of $512\times64\times64$. 

When we choose the \emph{no editing path}, we set $\lambda_{r1}=1.0$, $\lambda_{r2}=0.001$, $\lambda_{r3}=0.1$ for the face and $\lambda_{r3}=0.5$ for the car dataset. When we choose the \emph{cycle translation path}, we set $\lambda_{r1}=0.0$, meaning we do not use cycle consistency at the pixel level, $\lambda_{r2}=0.0001$,  $\lambda_{r3}=0.01$ for the face and $\lambda_{r3}=0.05$ for the car dataset. At both paths, we set $\lambda_{a}=0.1$. The regularizer coefficient is set to $\lambda_{f}=5.0$ for the face dataset and $\lambda_{f}=3.0$ for the car dataset.  
The network is trained with Adam optimizer, with a learning rate equal to $0.0001$. We halved the learning rate at iterations 5000, 10000 and 15000. 

\begin{figure}[h!]
\centering
\includegraphics[width=\linewidth]{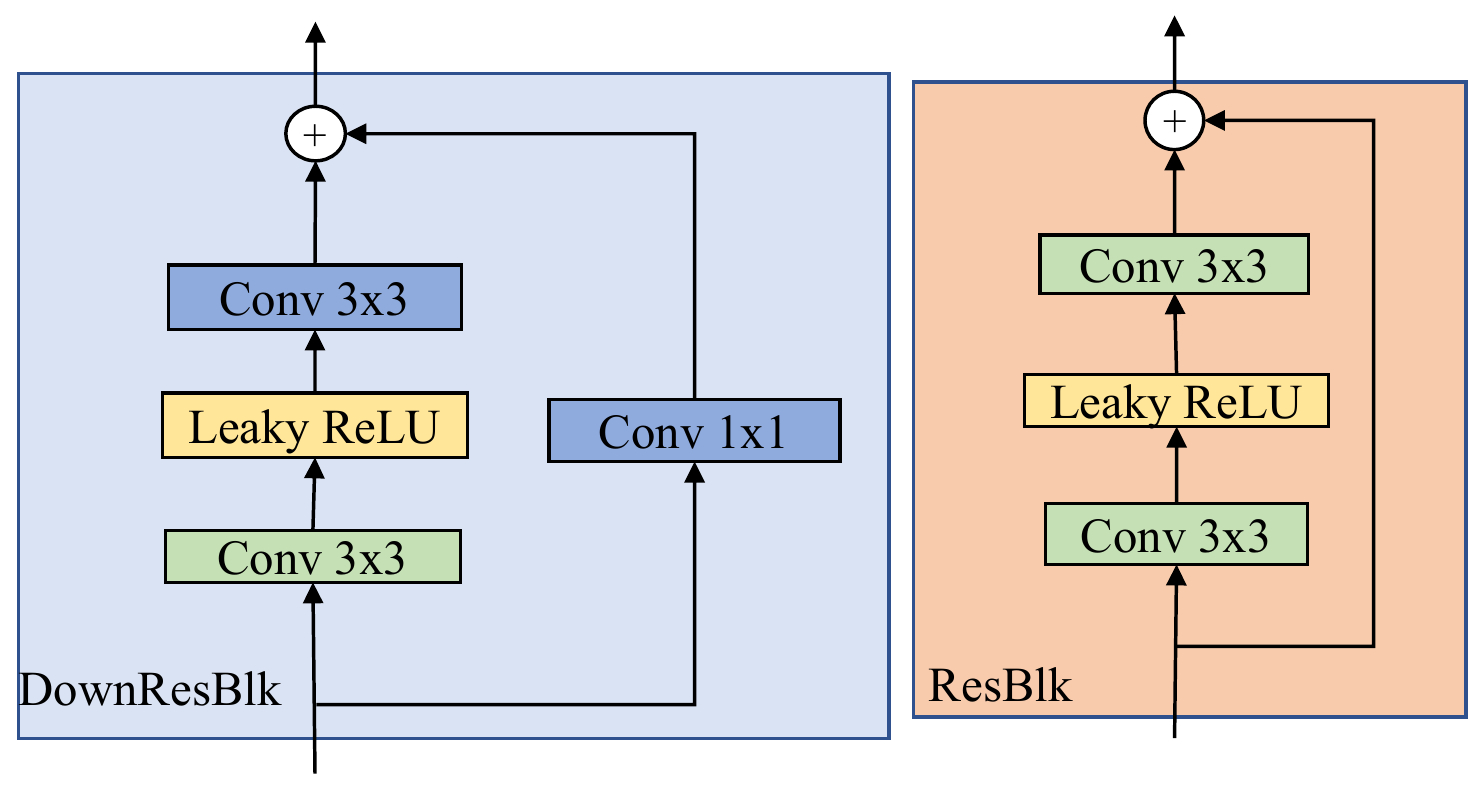}
\caption{Building blocks of our encoders. On the left, we demonstrate the downsampling layers used in $E_1$. On the right, we show standard residual layers used in both $E_1$ and $E_2$. }
\label{fig:res_layers}
\end{figure}

\section{Model Architecture}
Our model consists of residual layers, which are visualized in Fig. \ref{fig:res_layers}. Encoder $E_1$ takes the concatenated features $F_0$ and $G_{w}$, which has a spatial dimension of 640x64x64. It then forwards the concatenated features to $E_2$ using an encoder-decoder architecture. The first convolution layer sets the channel size as 512, and it is not changed in the rest of the layers. \emph{DownResBlk} reduces the resolution to half, and \emph{Interpolate} layer doubles the resolution, by using linear interpolation. 

$E_2$ takes both $F_a$ and $G_\alpha$ as inputs, and learns how to adapt encoded features to the generator features. Before concatenating the input features, their channel sizes are reduced to 256 using 3x3 convolution layers. After the concatenation, the channel size becomes 512, and it does not change throughout the $E_2$.  The detailed architecture of our encoders is given in Fig. \ref{fig:arch_details}.

\section{Evaluation Details}
For the face images, the reconstruction evaluations are performed on the last 2000 images of the CelebA-HQ dataset, same as the most of the inversion methods. 
For smile addition, we first find the ground truth smiling and non-smiling images in the CelebA test set. We add a smile to non-smiling ground truth images, and obtain fake smiling images. Finally, we compute the FID between the ground-truth smiling and the fake smiling images. Similar setup is used for smile removal as well. To add or remove a smile, we use the boundary obtained by InterfaceGAN, with a factor of 3.

For the car images, the reconstruction and editing evaluations are performed on the first 2000 images of the Stanford Cars test set. Because we do not have ground-truth labels, we directly calculate FID between the original and edited images. The edited images are obtained using GanSpace directions. 

For HyperStyle and Restyle, we use 5 iterative iterations in the reconstruction and editing, which is consistent with their training scheme. For PTI, we deploy locality regularization, as introduced in their work, for both reconstruction and editing.

We provide visuals obtained with various editing methods, namely InterfaceGan \cite{shen2020interpreting}, GanSpace \cite{harkonen2020ganspace}, StyleClip \cite{patashnik2021styleclip} and GradCtrl \cite{chen2022exploring}. Because our network is trained with editing in mind, we can directly apply boundaries found by different editing methods.
% This is also an advantage of our network when compared with inversion methods that can produce out-of-distribution inversions, like FeatureStyle. 

\section{Additional Results}
In Table \ref{table:results_pti}, we provide comparisons with PTI. 
PTI runs optimization for each image and achieves better reconstruction scores in terms of SSIM and LPIPS. 
However, our method achieves significantly better edit FIDs and reconstruction FIDs.
Furthermore, running evaluation with PTI method takes more than 2 days on a single GPU, whereas our method finishes within 5 minutes.
Run-time comparisons are provided in Table \ref{table:results_runtime}. The running times are obtained by averaging the reconstruction times of 2000 samples with batch size equal to 1 on a single NVIDIA GeForce GTX 1080 Ti GPU.  

\begin{figure*}
\centering
\scalebox{0.71}{
\addtolength{\tabcolsep}{-5pt}   
\begin{tabular}{ccccccccccc}
\rotatebox{90}{~~~~~~Smile (+)} &
\interpfigt{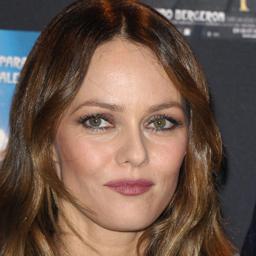} &
\interpfigt{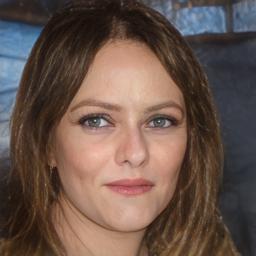} &
\interpfigt{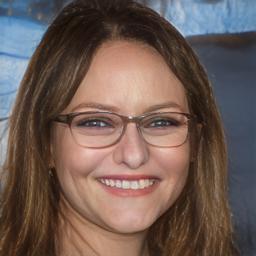} &
\interpfigt{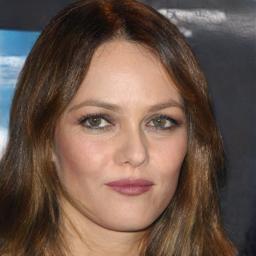} &
\interpfigt{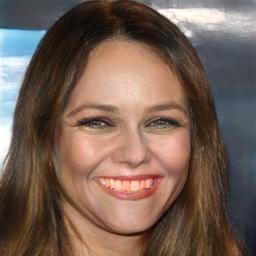} &
\interpfigt{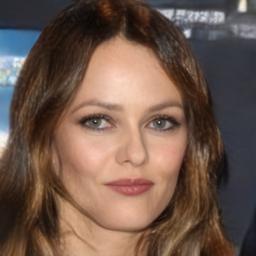} &
\interpfigt{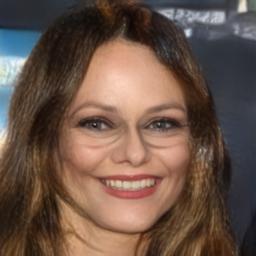} &
\interpfigt{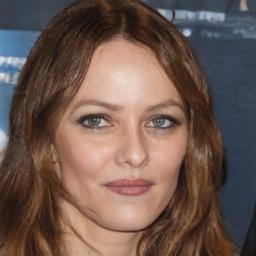} &
\interpfigt{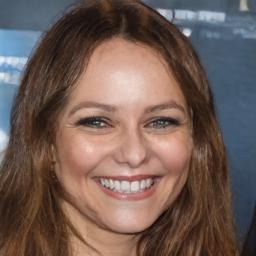}
\\
\rotatebox{90}{~~~~Smile (+)} &
\interpfigt{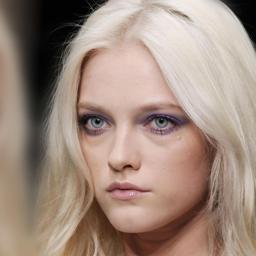} &
\interpfigt{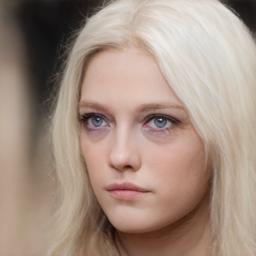} &
\interpfigt{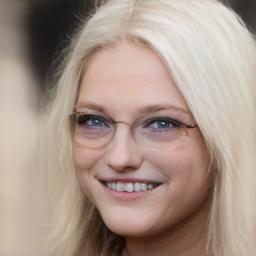} &
\interpfigt{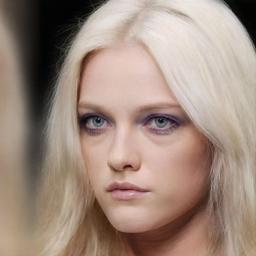} &
\interpfigt{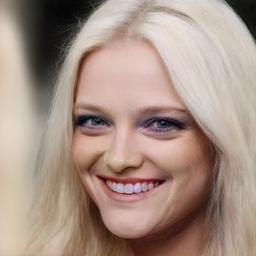} &
\interpfigt{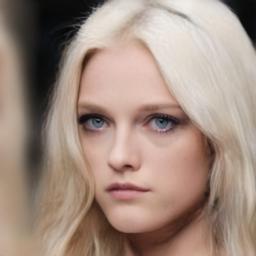} &
\interpfigt{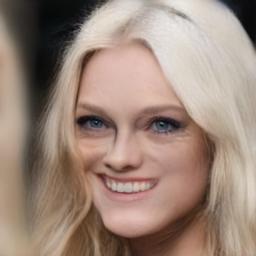} &
\interpfigt{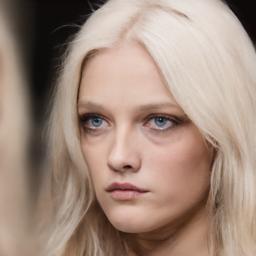} &
\interpfigt{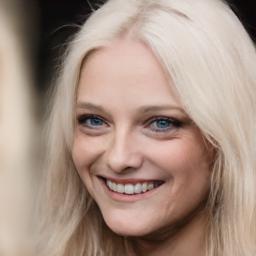}
\\
\rotatebox{90}{~~~~~~Smile (+)} &
\interpfigt{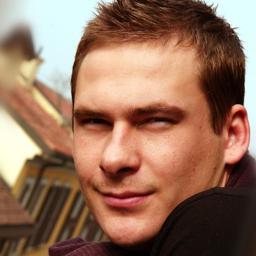} &
\interpfigt{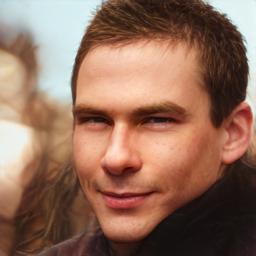} &
\interpfigt{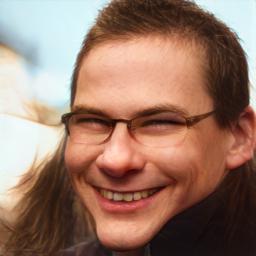} &
\interpfigt{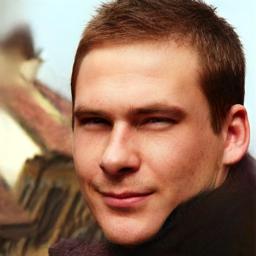} &
\interpfigt{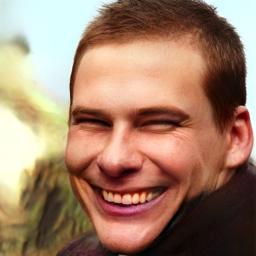} &
\interpfigt{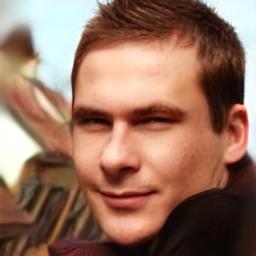} &
\interpfigt{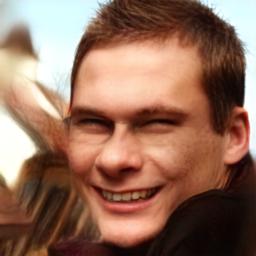} &
\interpfigt{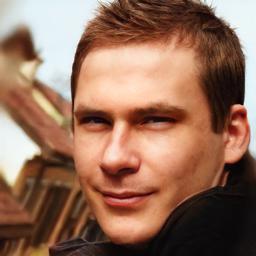} &
\interpfigt{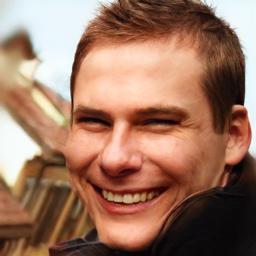}
\\
\rotatebox{90}{~~~~~~Smile (+)} &
\interpfigt{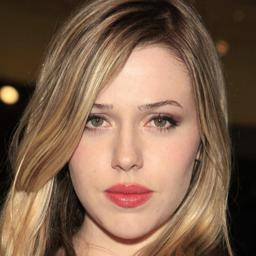} &
\interpfigt{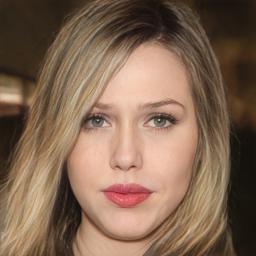} &
\interpfigt{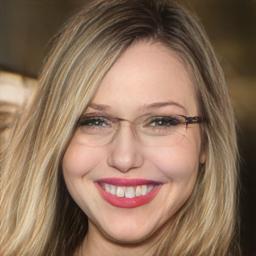} &
\interpfigt{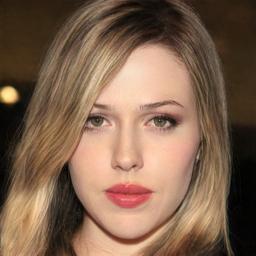} &
\interpfigt{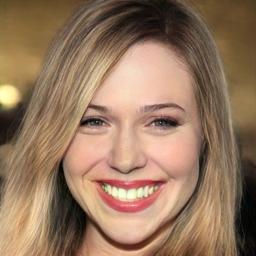} &
\interpfigt{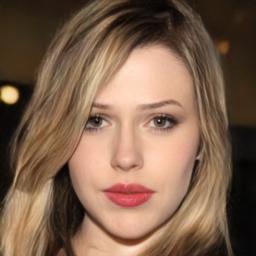} &
\interpfigt{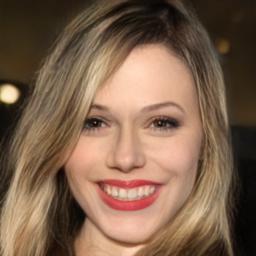} &
\interpfigt{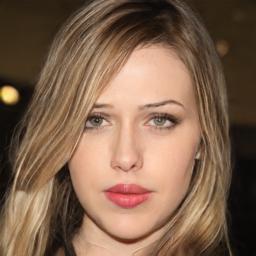} &
\interpfigt{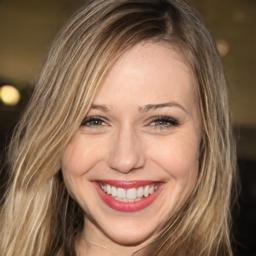}
\\
\rotatebox{90}{~~~~~~Smile (+)} &
\interpfigt{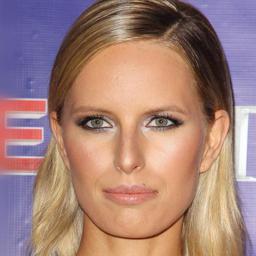} &
\interpfigt{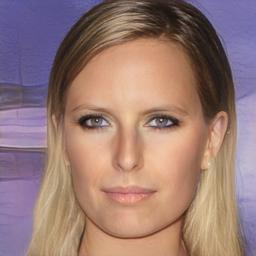} &
\interpfigt{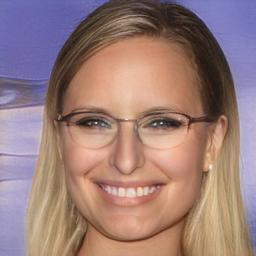} &
\interpfigt{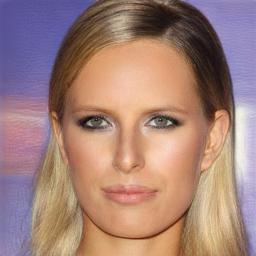} &
\interpfigt{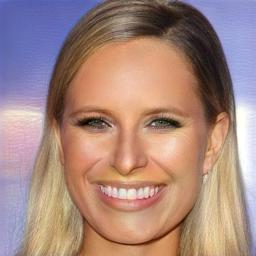} &
\interpfigt{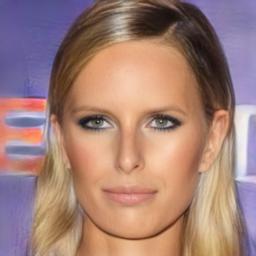} &
\interpfigt{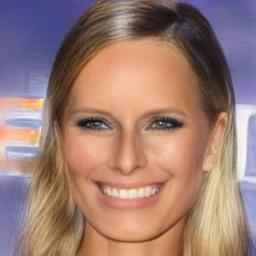} &
\interpfigt{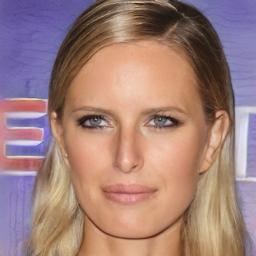} &
\interpfigt{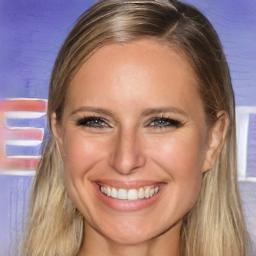}
\\
\rotatebox{90}{~~~~~~Smile (-)} &
\interpfigt{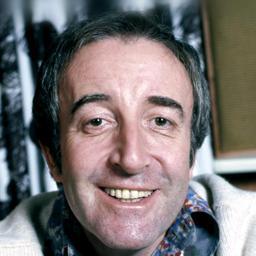} &
\interpfigt{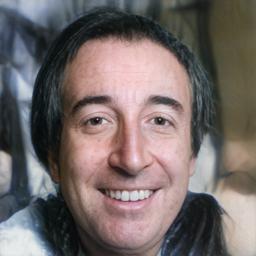} &
\interpfigt{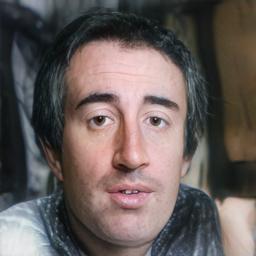} &
\interpfigt{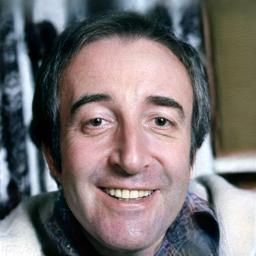} &
\interpfigt{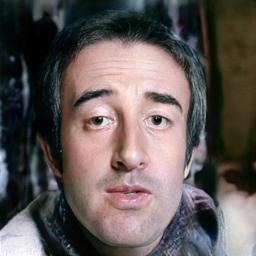} &
\interpfigt{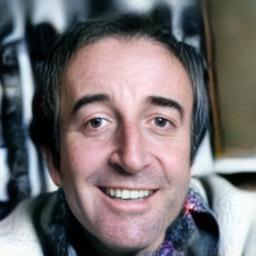} &
\interpfigt{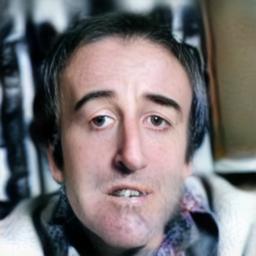} &
\interpfigt{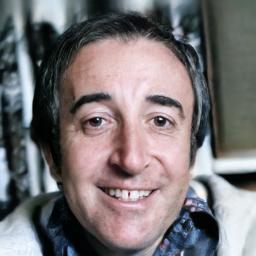} &
\interpfigt{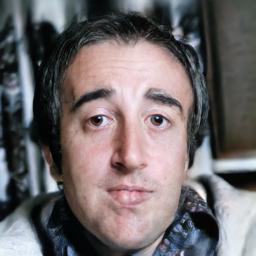}
\\
\rotatebox{90}{~~~~~~Smile (-)} &
\interpfigt{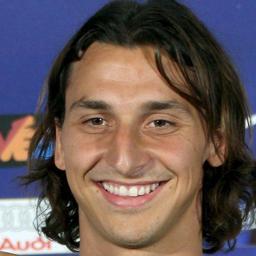} &
\interpfigt{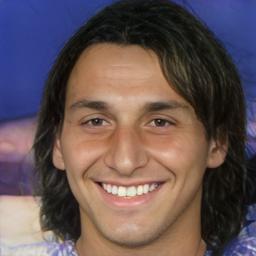} &
\interpfigt{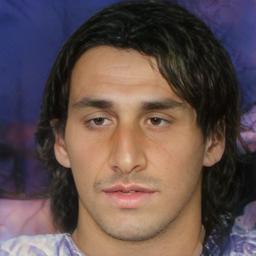} &
\interpfigt{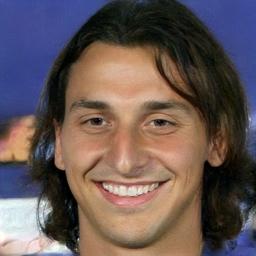} &
\interpfigt{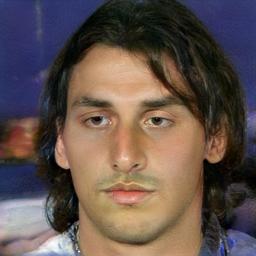} &
\interpfigt{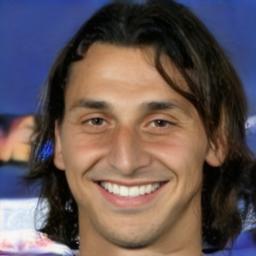} &
\interpfigt{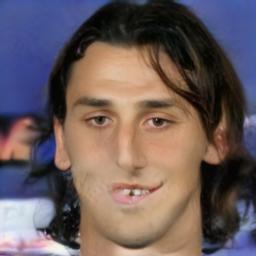} &
\interpfigt{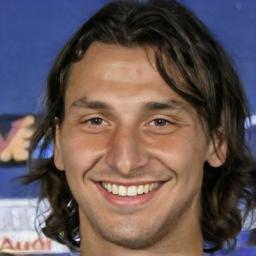} &
\interpfigt{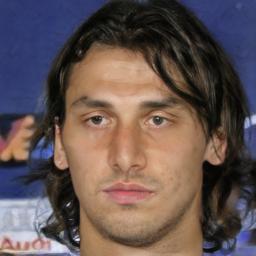} 
\\
\rotatebox{90}{~~~~~~Smile (-)} &
\interpfigt{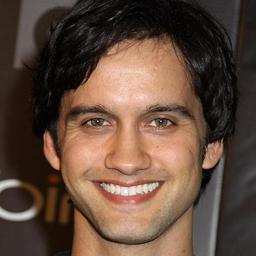} &
\interpfigt{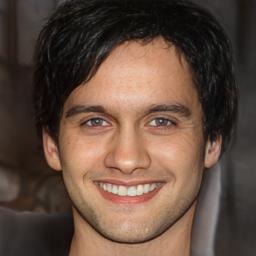} &
\interpfigt{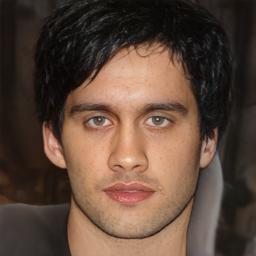} &
\interpfigt{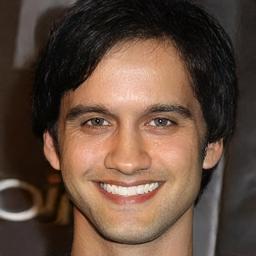} &
\interpfigt{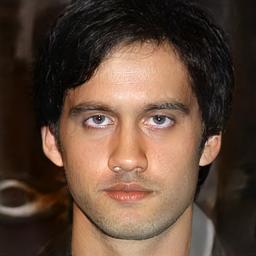} &
\interpfigt{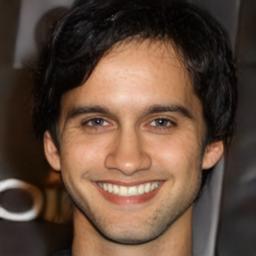} &
\interpfigt{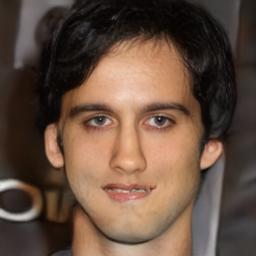} &
\interpfigt{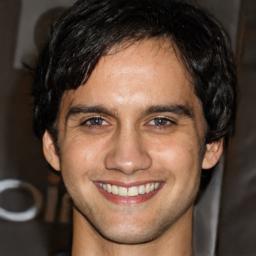} &
\interpfigt{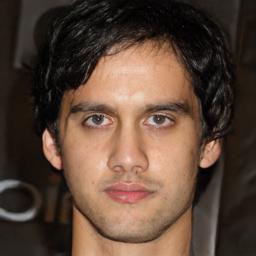} 
\\
\rotatebox{90}{~~~~~~Smile (-)} &
\interpfigt{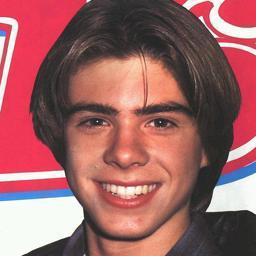} &
\interpfigt{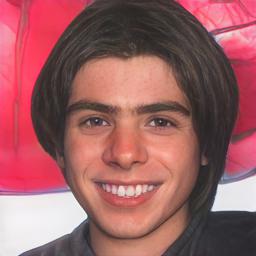} &
\interpfigt{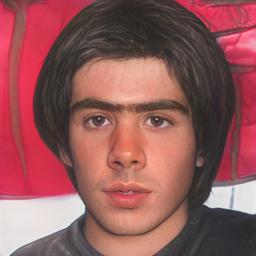} &
\interpfigt{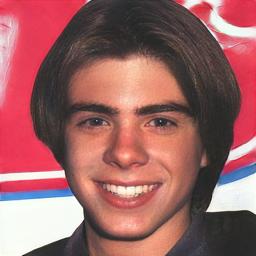} &
\interpfigt{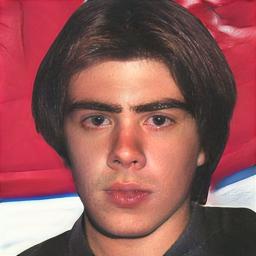} &
\interpfigt{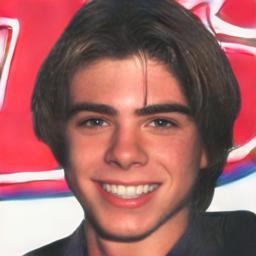} &
\interpfigt{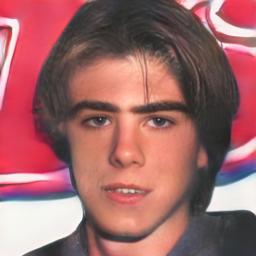} &
\interpfigt{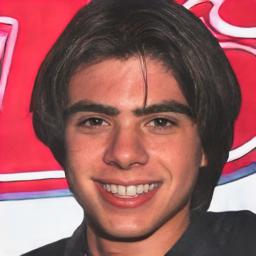} &
\interpfigt{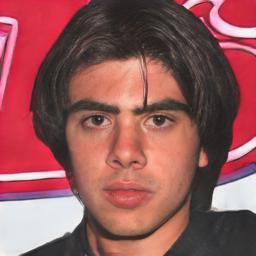} 
\\
\rotatebox{90}{~~~~~~Smile (-)} &
\interpfigt{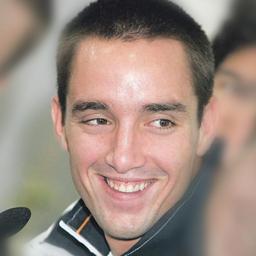} &
\interpfigt{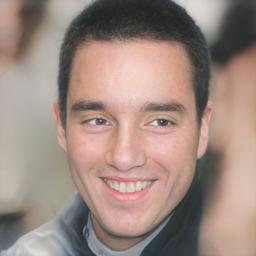} &
\interpfigt{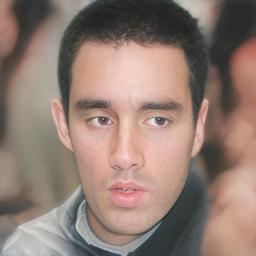} &
\interpfigt{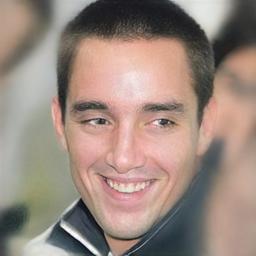} &
\interpfigt{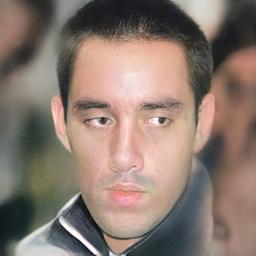} &
\interpfigt{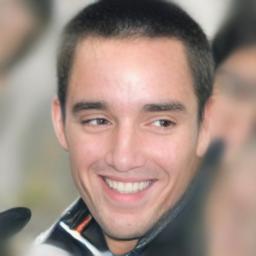} &
\interpfigt{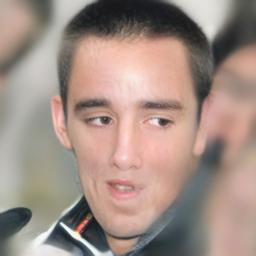} &
\interpfigt{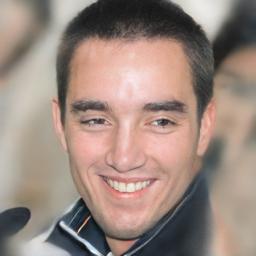} &
\interpfigt{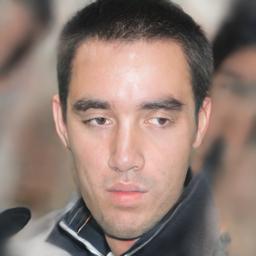} 
\\
&  Input &  \multicolumn{2}{c}{StyleTransformer \cite{Hu_2022_CVPR}} & \multicolumn{2}{c}{PTI \cite{roich2022pivotal}} &  \multicolumn{2}{c}{FeatureStyle \cite{xuyao2022}} &  \multicolumn{2}{c}{StyleRes (Ours)} \\

\end{tabular}
}
\caption{Qualitative results of inversion and editing. For each method, first column shows inversion, and second shows editing.}
\label{fig:results_smile_add_others}
\end{figure*}

\begin{figure*}
\centering
\scalebox{0.71}{
\addtolength{\tabcolsep}{-5pt}   
\begin{tabular}{ccccccccccc}
\rotatebox{90}{~~~~~~Smile (+)} &
\interpfigt{Figures/face_imgs/smile_add/input/26.jpg} &
\interpfigt{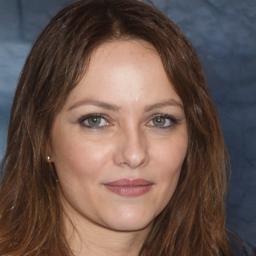} &
\interpfigt{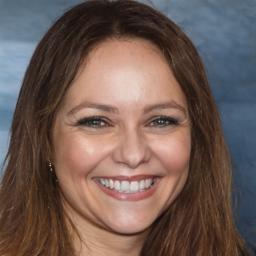} &
\interpfigt{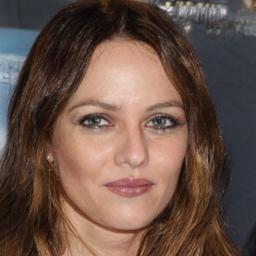} &
\interpfigt{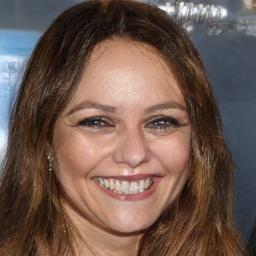} &
\interpfigt{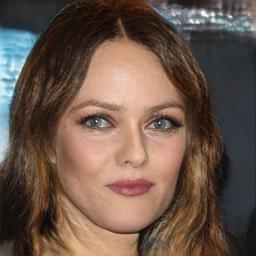} &
\interpfigt{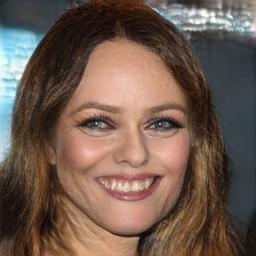} &
\interpfigt{Figures/face_imgs/inversion/ours/26.jpg} &
\interpfigt{Figures/face_imgs/smile_add/ours/26.jpg}
\\ 
\rotatebox{90}{~~~~~~~Age (-)} &
\interpfigt{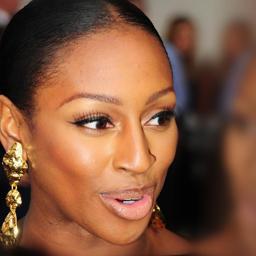} &
\interpfigt{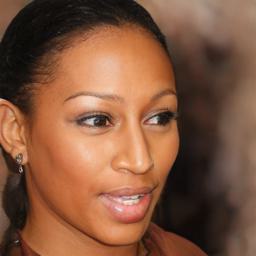} &
\interpfigt{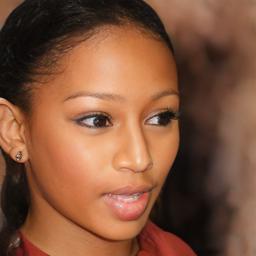} &
\interpfigt{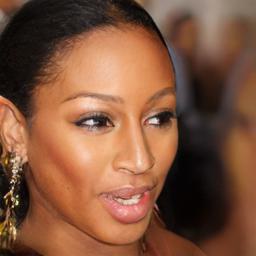} &
\interpfigt{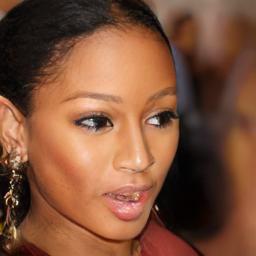} &
\interpfigt{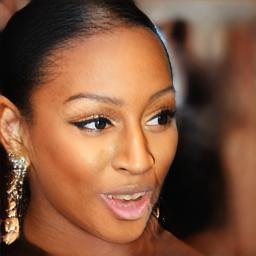} &
\interpfigt{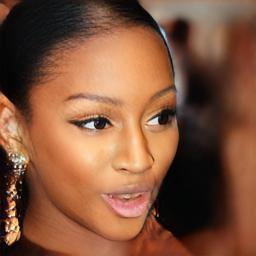} &
\interpfigt{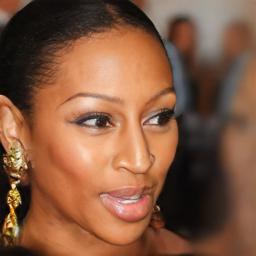} &
\interpfigt{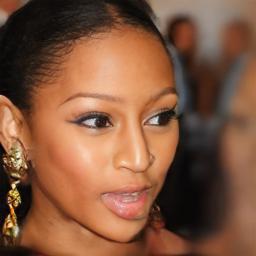}
\\
\rotatebox{90}{~~~~~~~Age (+)} &
\interpfigt{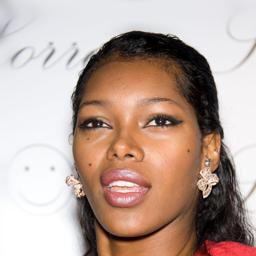} &
\interpfigt{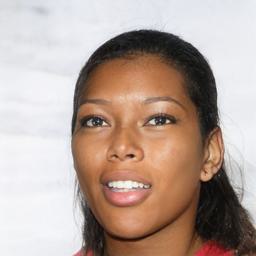} &
\interpfigt{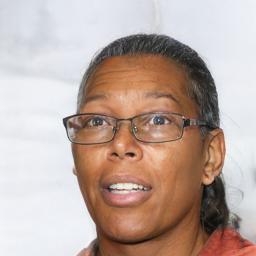} &
\interpfigt{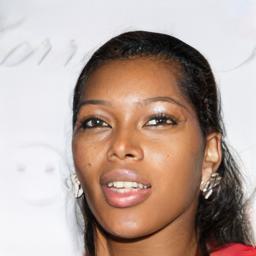} &
\interpfigt{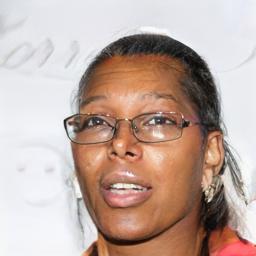} &
\interpfigt{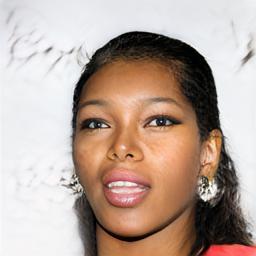} &
\interpfigt{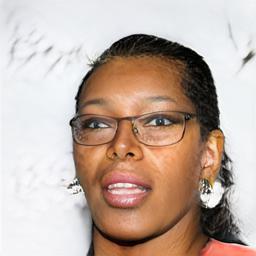} &
\interpfigt{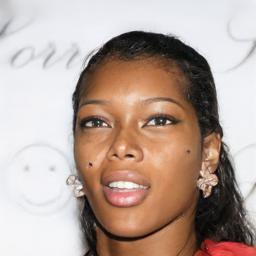} &
\interpfigt{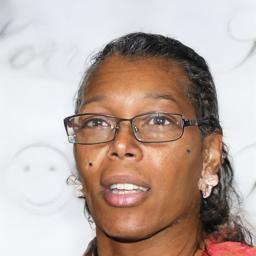}
\\
\rotatebox{90}{~~~~~~~Pose (+)} &
\interpfigt{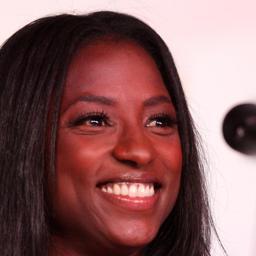} &
\interpfigt{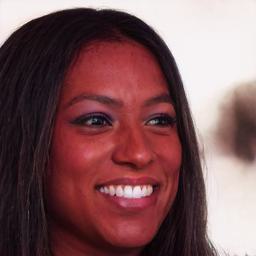} &
\interpfigt{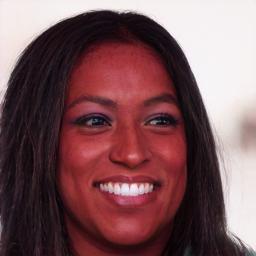} &
\interpfigt{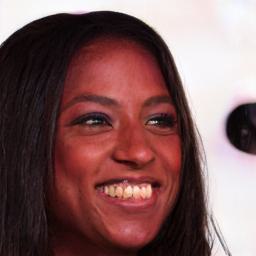} &
\interpfigt{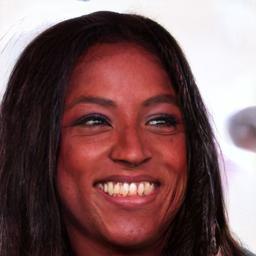} &
\interpfigt{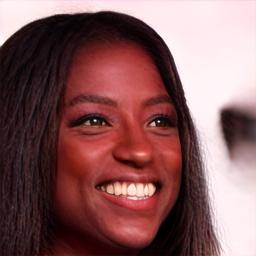} &
\interpfigt{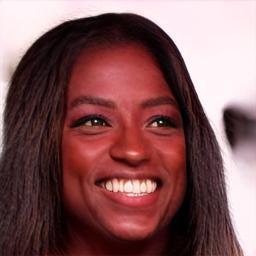} &
\interpfigt{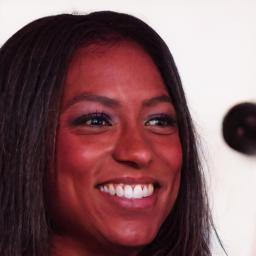} &
\interpfigt{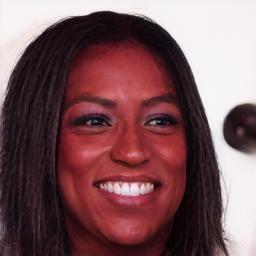} 
\\
\rotatebox{90}{~~~~~~~Pose (-)} &
\interpfigt{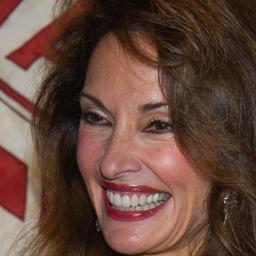} &
\interpfigt{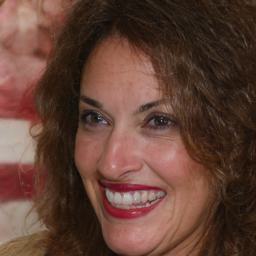} &
\interpfigt{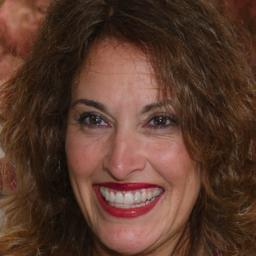} &
\interpfigt{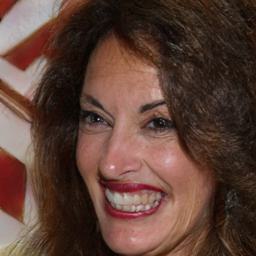} &
\interpfigt{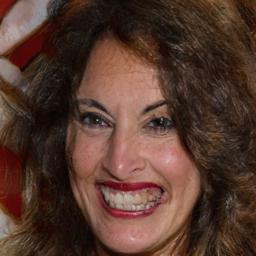} &
\interpfigt{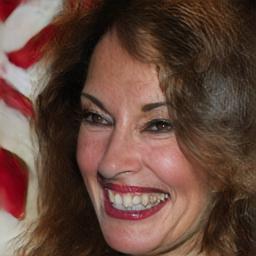} &
\interpfigt{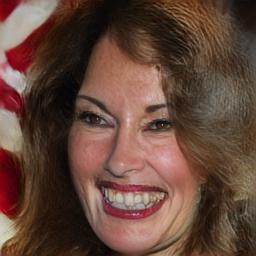} &
\interpfigt{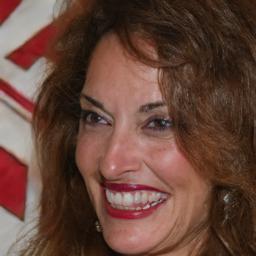} &
\interpfigt{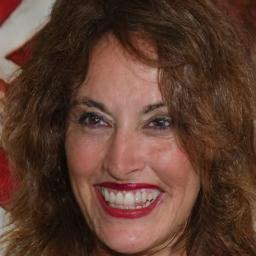}
\\
\rotatebox{90}{~~~~~~~~~Beard} &
\interpfigt{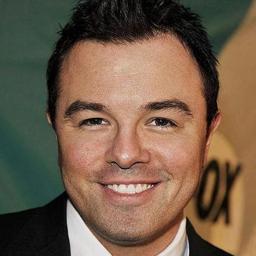} &
\interpfigt{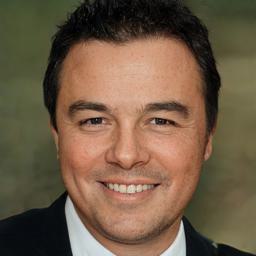} &
\interpfigt{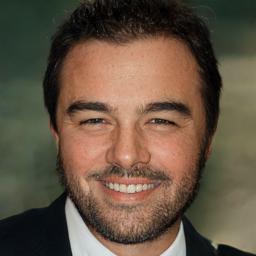} &
\interpfigt{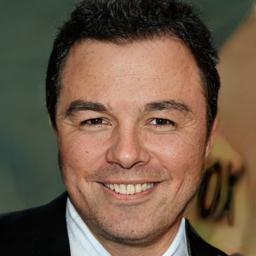} &
\interpfigt{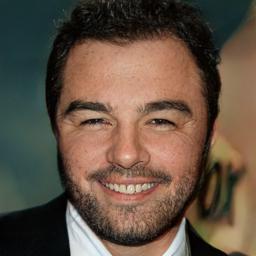} &
\interpfigt{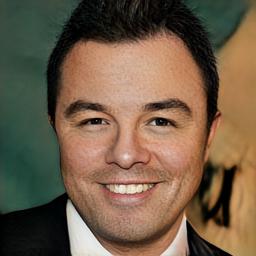} &
\interpfigt{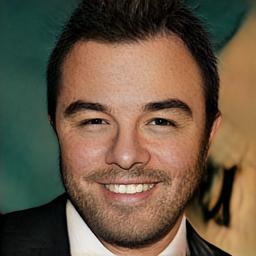} &
\interpfigt{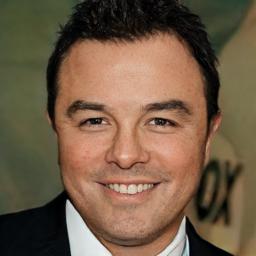} &
\interpfigt{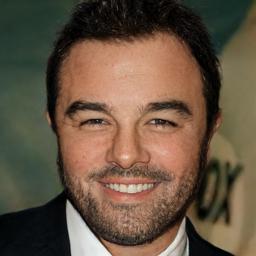}
\\
\rotatebox{90}{~~~Eye Openness} &
\interpfigt{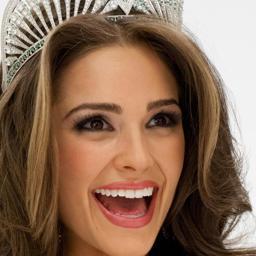} &
\interpfigt{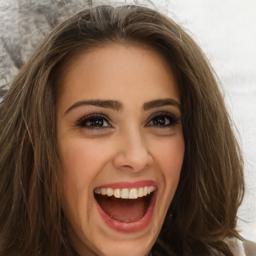} &
\interpfigt{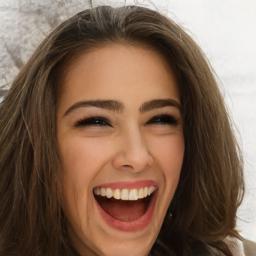} &
\interpfigt{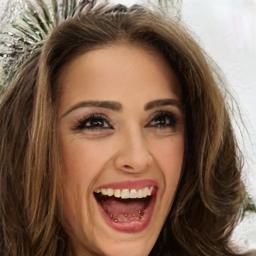} &
\interpfigt{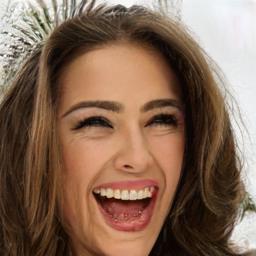} &
\interpfigt{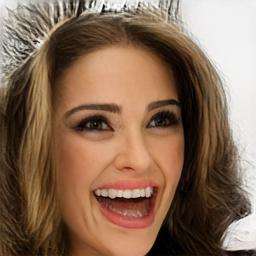} &
\interpfigt{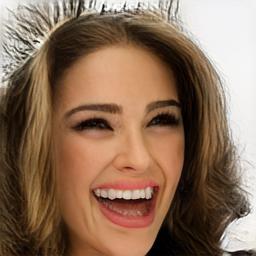} &
\interpfigt{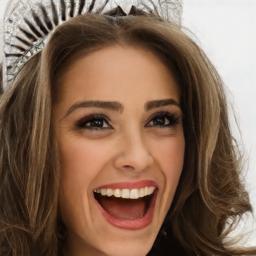} &
\interpfigt{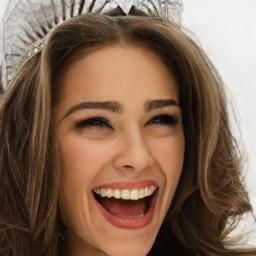}
\\

\rotatebox{90}{~~~~~~Lipstick} &1
\interpfigt{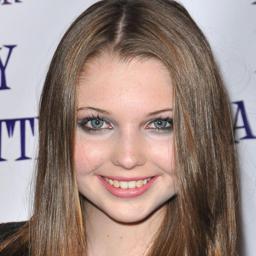} &
\interpfigt{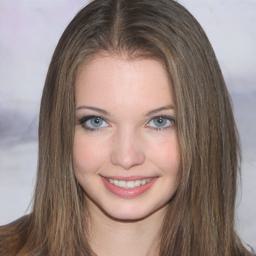} &
\interpfigt{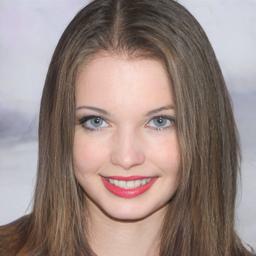} &
\interpfigt{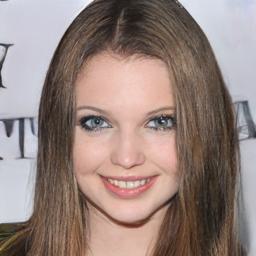} &
\interpfigt{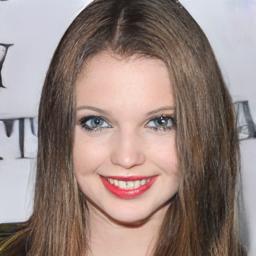} &
\interpfigt{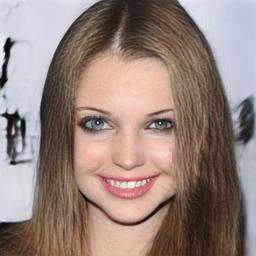} &
\interpfigt{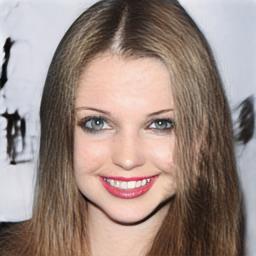} &
\interpfigt{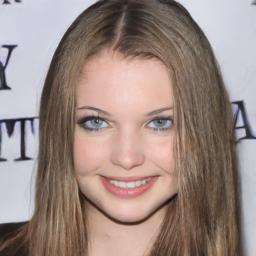} &
\interpfigt{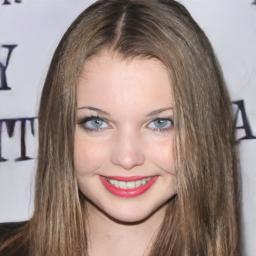}
\\
\rotatebox{90}{~~~~~~Eyeglasses} &
\interpfigt{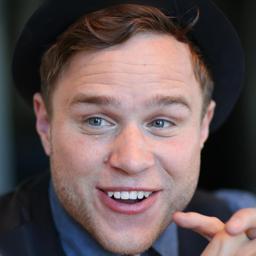} &
\interpfigt{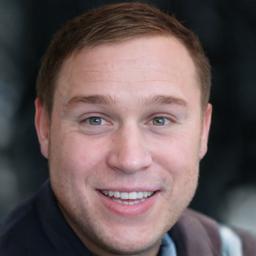} &
\interpfigt{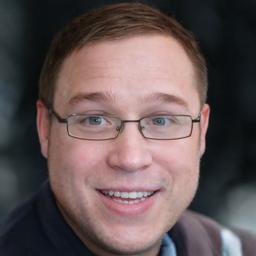} &
\interpfigt{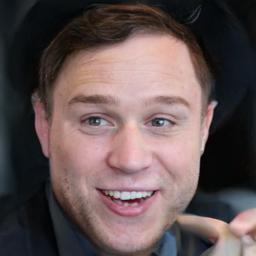} &
\interpfigt{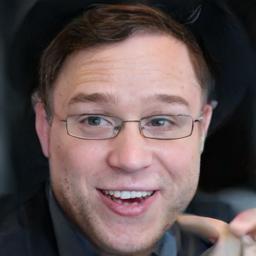} &
\interpfigt{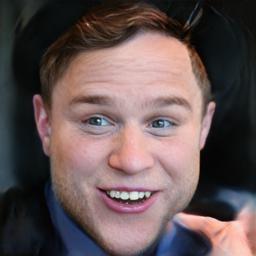} &
\interpfigt{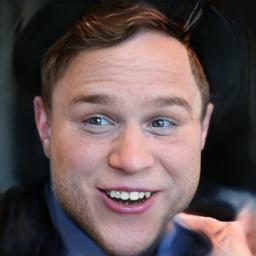} &
\interpfigt{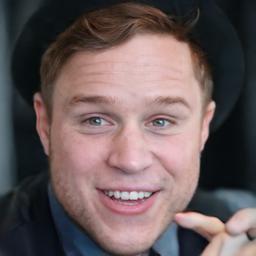} &
\interpfigt{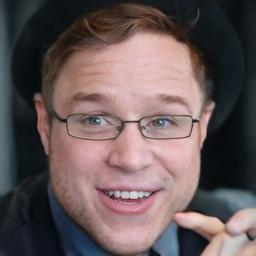}
\\
\rotatebox{90}{~~~~~~~~~Bangs} &
\interpfigt{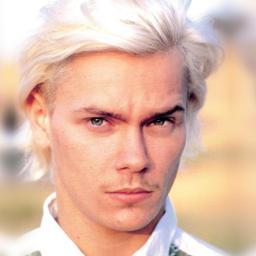} &
\interpfigt{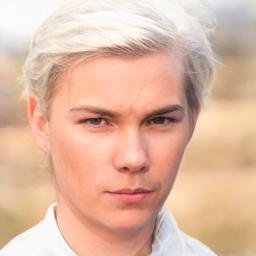} &
\interpfigt{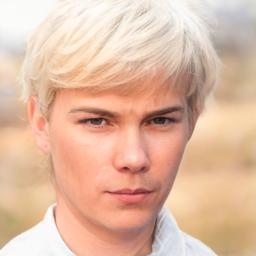} &
\interpfigt{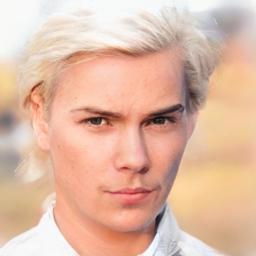} &
\interpfigt{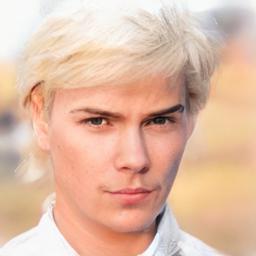} &
\interpfigt{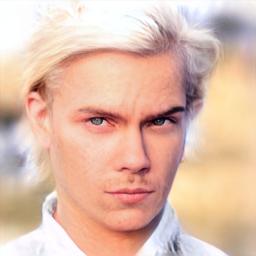} &
\interpfigt{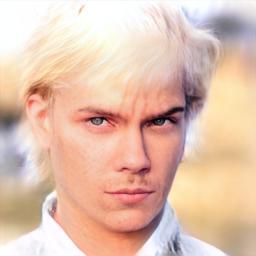} &
\interpfigt{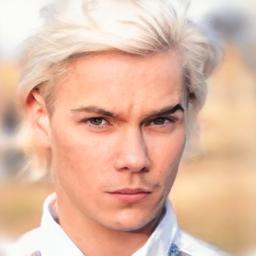} &
\interpfigt{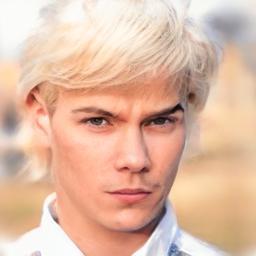}
\\
\rotatebox{90}{~~~~~~~Bob Cut} &
\interpfigt{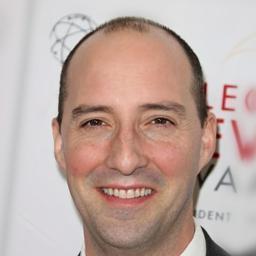} &
\interpfigt{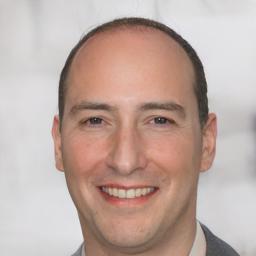} &
\interpfigt{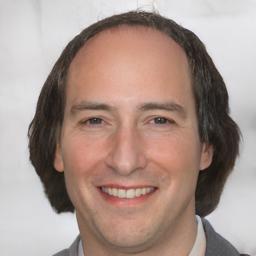} &
\interpfigt{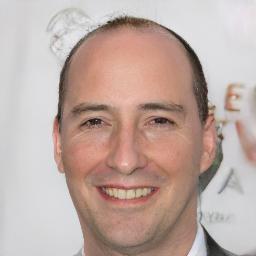} &
\interpfigt{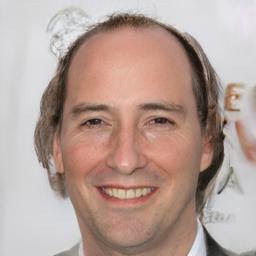} &
\interpfigt{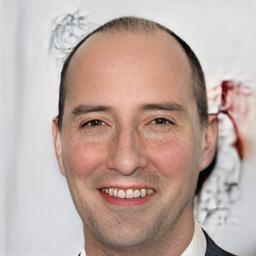} &
\interpfigt{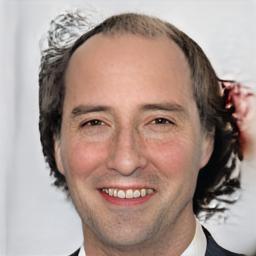} &
\interpfigt{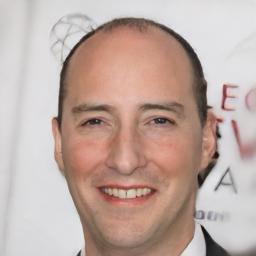} &
\interpfigt{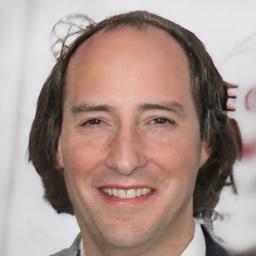}
\\

&  Input &  \multicolumn{2}{c}{e4e \cite{tov2021designing}} &  \multicolumn{2}{c}{HFGI \cite{wang2022high}} &  \multicolumn{2}{c}{HyperStyle \cite{alaluf2022hyperstyle}}  & \multicolumn{2}{c}{StyleRes (Ours)} \\

\end{tabular}
}
\caption{Qualitative results of inversion and editing. For each method, first column shows inversion, and second shows editing.}
\label{fig:results_smile_add}
\end{figure*}

\begin{figure*}
\centering
\scalebox{0.71}{
\addtolength{\tabcolsep}{-5pt}   
\begin{tabular}{ccccccccccc}
\rotatebox{90}{~~Car Color} &
\interpfigt{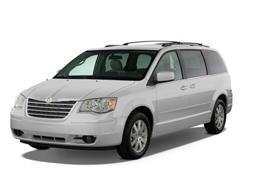} &
\interpfigt{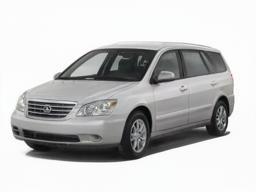} &
\interpfigt{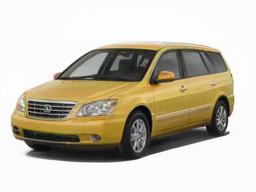} &
\interpfigt{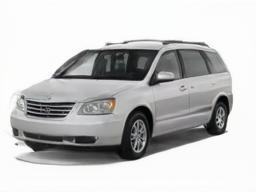} &
\interpfigt{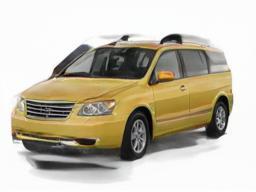} &
\interpfigt{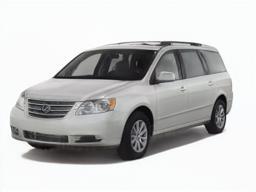} &
\interpfigt{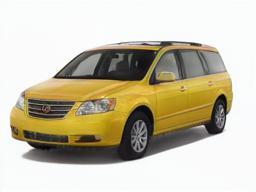} &
\interpfigt{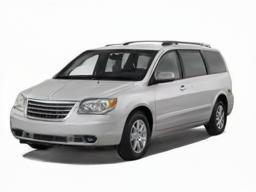} &
\interpfigt{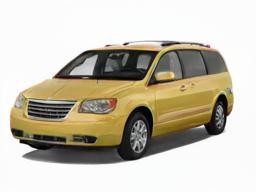}
\\
\rotatebox{90}{~~Car Color} &
\interpfigt{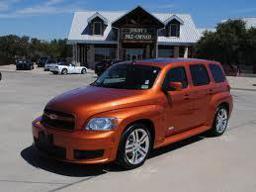} &
\interpfigt{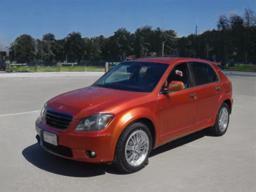} &
\interpfigt{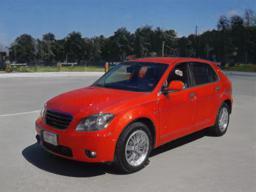} &
\interpfigt{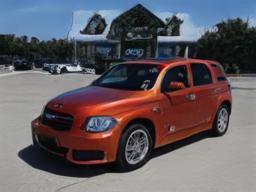} &
\interpfigt{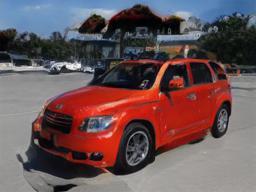} &
\interpfigt{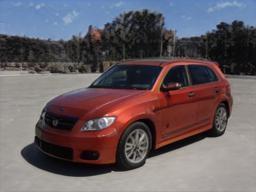} &
\interpfigt{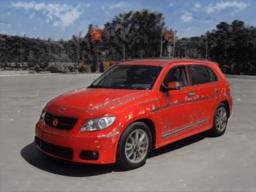} &
\interpfigt{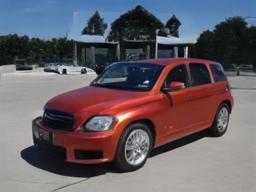} &
\interpfigt{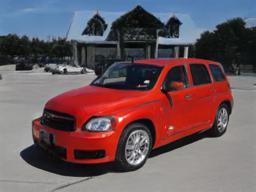}
\\
\rotatebox{90}{~~Car Color} &
\interpfigt{Figures/car_imgs/color/input/00024.jpg} &
\interpfigt{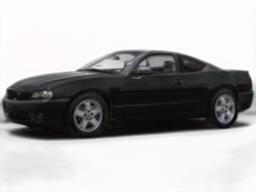} &
\interpfigt{Figures/car_imgs/color/e4e/00024.jpg} &
\interpfigt{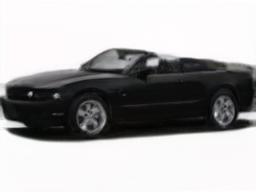} &
\interpfigt{Figures/car_imgs/color/hfgi/00024.jpg} &
\interpfigt{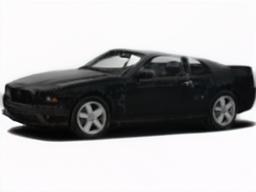} &
\interpfigt{Figures/car_imgs/color/hyperstyle/00024.jpg} &
\interpfigt{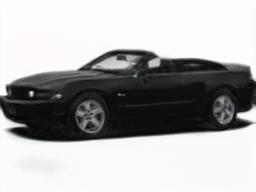} &
\interpfigt{Figures/car_imgs/color/ours/00024.jpg}
\\
\rotatebox{90}{~~Car Color} &
\interpfigt{Figures/car_imgs/color/input/00037.jpg} &
\interpfigt{Figures/car_imgs/inversion/e4e/00037.jpg} &
\interpfigt{Figures/car_imgs/color/e4e/00037.jpg} &
\interpfigt{Figures/car_imgs/inversion/hfgi/00037.jpg} &
\interpfigt{Figures/car_imgs/color/hfgi/00037.jpg} &
\interpfigt{Figures/car_imgs/inversion/hyperstyle/00037.jpg} &
\interpfigt{Figures/car_imgs/color/hyperstyle/00037.jpg} &
\interpfigt{Figures/car_imgs/inversion/ours/00037.jpg} &
\interpfigt{Figures/car_imgs/color/ours/00037.jpg}
\\
\rotatebox{90}{~~Car Color} &
\interpfigt{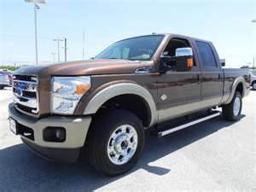} &
\interpfigt{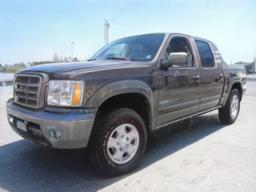} &
\interpfigt{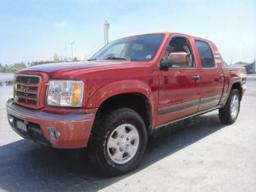} &
\interpfigt{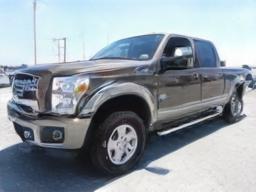} &
\interpfigt{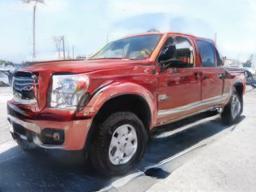} &
\interpfigt{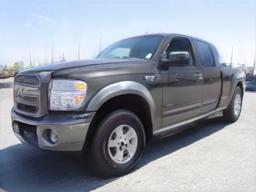} &
\interpfigt{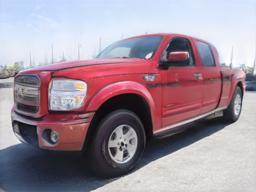} &
\interpfigt{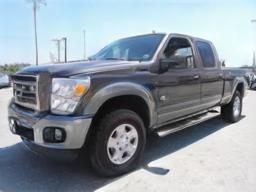} &
\interpfigt{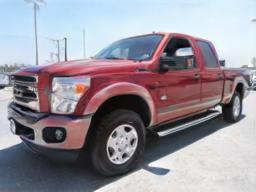}
\\
\rotatebox{90}{~~Car Color} &
\interpfigt{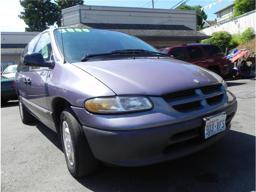} &
\interpfigt{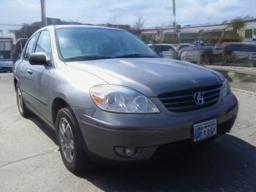} &
\interpfigt{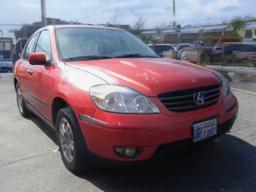} &
\interpfigt{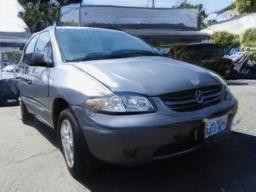} &
\interpfigt{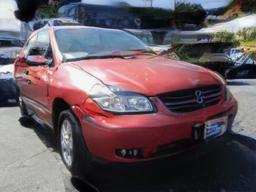} &
\interpfigt{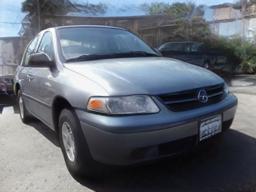} &
\interpfigt{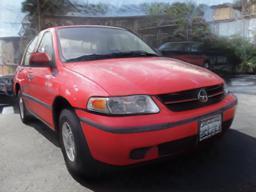} &
\interpfigt{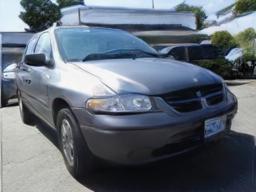} &
\interpfigt{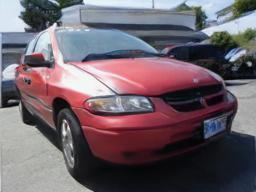}
\\
\rotatebox{90}{~~~~~~Grass} &
\interpfigt{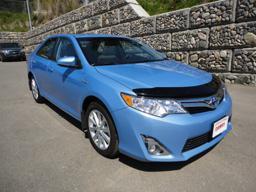} &
\interpfigt{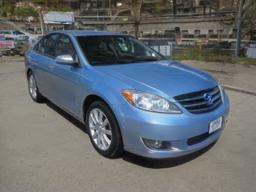} &
\interpfigt{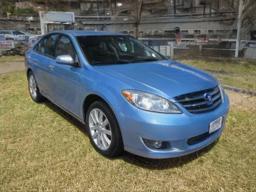} &
\interpfigt{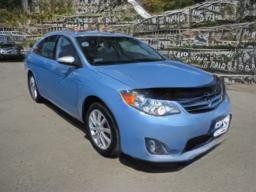} &
\interpfigt{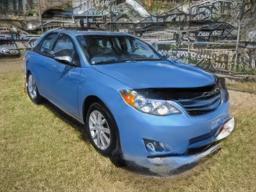} &
\interpfigt{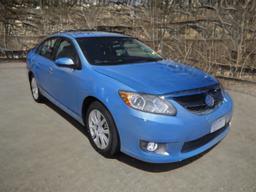} &
\interpfigt{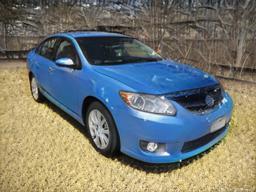} &
\interpfigt{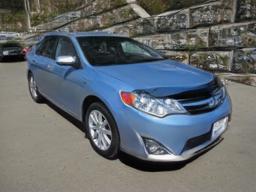} &
\interpfigt{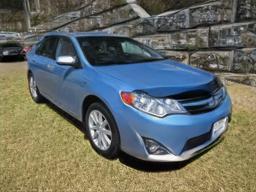}
\\
\rotatebox{90}{~~~~~~Grass} &
\interpfigt{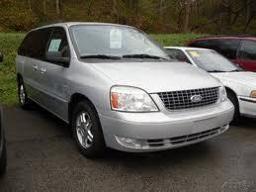} &
\interpfigt{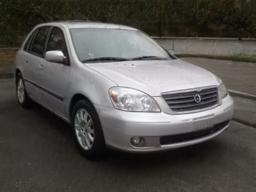} &
\interpfigt{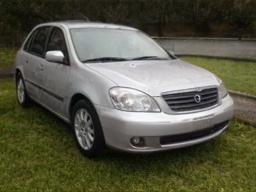} &
\interpfigt{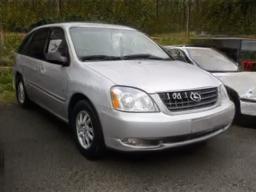} &
\interpfigt{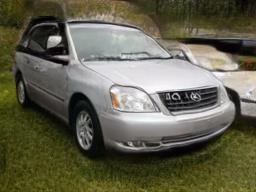} &
\interpfigt{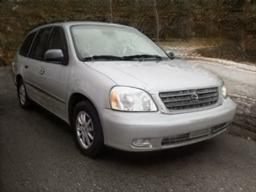} &
\interpfigt{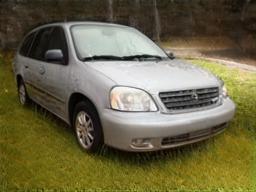} &
\interpfigt{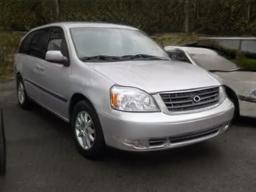} &
\interpfigt{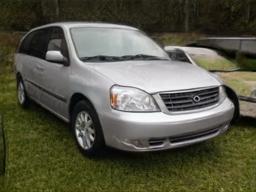}
\\
\rotatebox{90}{~~~~~~Grass} &
\interpfigt{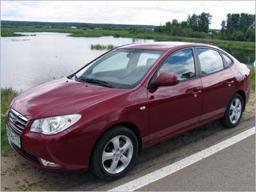} &
\interpfigt{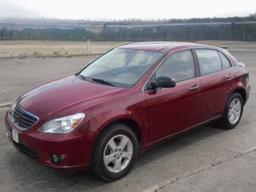} &
\interpfigt{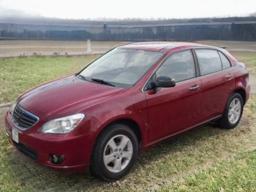} &
\interpfigt{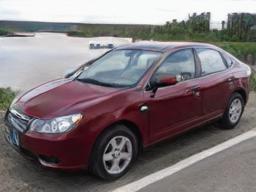} &
\interpfigt{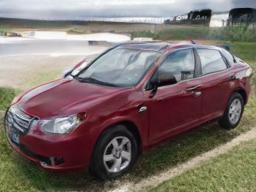} &
\interpfigt{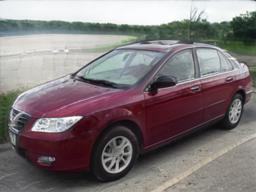} &
\interpfigt{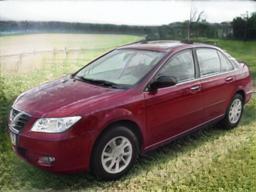} &
\interpfigt{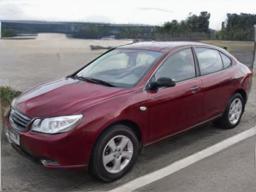} &
\interpfigt{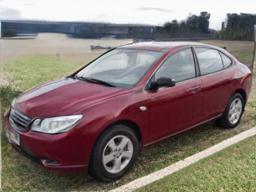}
\\
\rotatebox{90}{~~~~~~Grass} &
\interpfigt{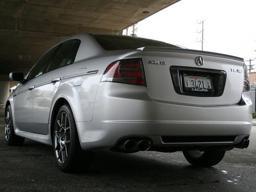} &
\interpfigt{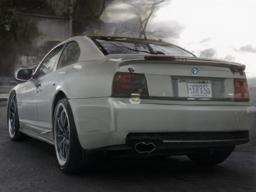} &
\interpfigt{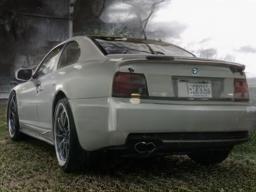} &
\interpfigt{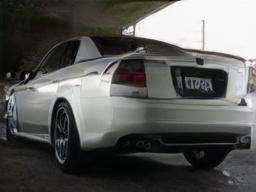} &
\interpfigt{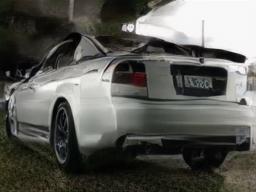} &
\interpfigt{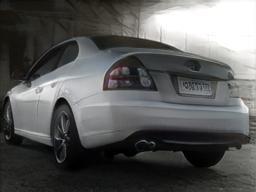} &
\interpfigt{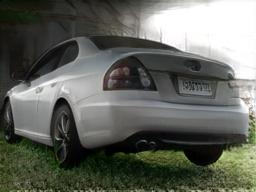} &
\interpfigt{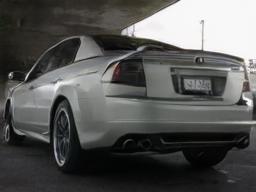} &
\interpfigt{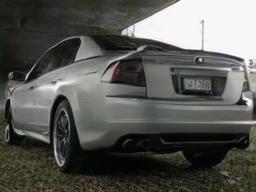}
\\
\rotatebox{90}{~~~~~~Grass} &
\interpfigt{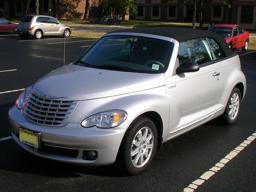} &
\interpfigt{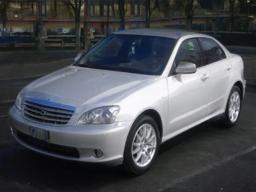} &
\interpfigt{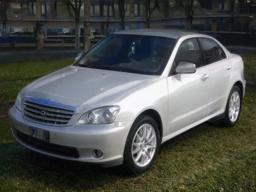} &
\interpfigt{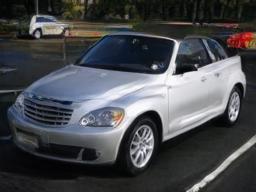} &
\interpfigt{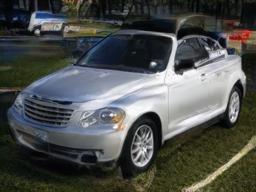} &
\interpfigt{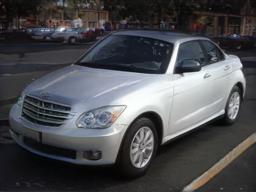} &
\interpfigt{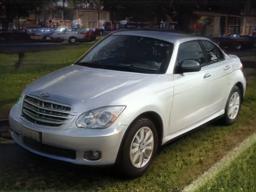} &
\interpfigt{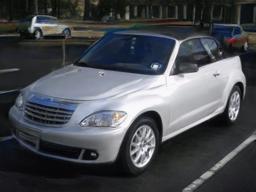} &
\interpfigt{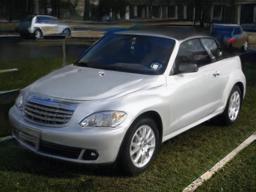}
\\
\rotatebox{90}{~~~~~~Grass} &
\interpfigt{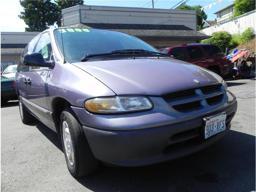} &
\interpfigt{Figures/car_imgs/inversion/e4e/00048.jpg} &
\interpfigt{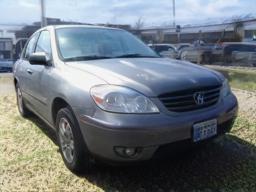} &
\interpfigt{Figures/car_imgs/inversion/hfgi/00048.jpg} &
\interpfigt{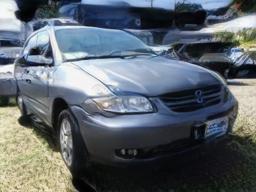} &
\interpfigt{Figures/car_imgs/inversion/hyperstyle/00048.jpg} &
\interpfigt{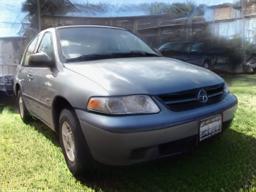} &
\interpfigt{Figures/car_imgs/inversion/ours/00048.jpg} &
\interpfigt{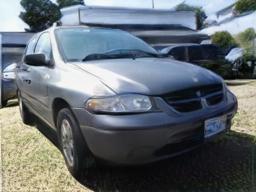}
\\

&  Input &  \multicolumn{2}{c}{e4e \cite{tov2021designing}} &  \multicolumn{2}{c}{HFGI \cite{wang2022high}} &  \multicolumn{2}{c}{HyperStyle \cite{alaluf2022hyperstyle}}  & \multicolumn{2}{c}{StyleRes (Ours)} \\
\end{tabular}
}
\caption{Qualitative results of inversion and editing. For each method, first column shows inversion, and second shows editing.}
\label{fig:results_car_color}
\end{figure*}

We provide visual comparisons on:
\begin{enumerate}
    \item smile editing with StyleTransformer, PTI, and FeatureStyle in Figs. \ref{fig:results_smile_add_others}, by using InterfaceGAN.
     \item various editings with e4e, HFGI and HyperStyle in Figs. \ref{fig:results_smile_add}, by using InterfaceGAN, GanSpace and StyleClip. 
    \item car editings with e4e, HFGI and  HyperStyle in Fig. \ref{fig:results_car_color}, by using GanSpace.
\end{enumerate}

